\pdfoutput=1

\documentclass[11pt]{article}

\usepackage[acceptedWithA]{tacl2021v1}
\usepackage{times}
\usepackage{latexsym}

\usepackage[T1]{fontenc}

\usepackage{microtype}

\usepackage{inconsolata}

\usepackage{graphicx}

\usepackage{xcolor}
\usepackage{booktabs}
\usepackage{xspace}
\usepackage{multirow}
\usepackage[normalem]{ulem}
\usepackage{booktabs,subcaption,amsfonts,dcolumn}

\usepackage{tcolorbox}
\usepackage{tabularx}

\usepackage{todonotes}

\makeatletter
\newcommand*\iftodonotes{\if@todonotes@disabled\expandafter\@secondoftwo\else\expandafter\@firstoftwo\fi}
\makeatother

\newcommand{\note}[4][]{\todo[author=#2,color=#3,size=\scriptsize,fancyline,caption={},#1]{#4}}

\definecolor{ivanlime}{rgb}{0.9,1,0.3}
\definecolor{awesomered}{rgb}{1.0, 0.13, 0.32}

\newcommand{\hannah}[2][]{\note[#1]{hannah}{white}{#2}}

\newcommand{\rparagraph}[1]{\vspace{1.2mm}\noindent\textbf{#1.}}

\newcommand{\sparagraph}[1]{\vspace{0.0mm}\noindent\textbf{#1.}}

\newcommand{\rparagraphnodot}[1]{\vspace{1.2mm}\noindent\textbf{#1}}

\newcommand{\rVQA}[0]{\textsc{DARE}\xspace}

\definecolor{darkgreen}{rgb}{0.024, 0.631, 0.094}
\newcommand{\edit}[1]{\textcolor{darkgreen}{#1}}

\newif\iftaclinstructions
\taclinstructionsfalse 
\iftaclinstructions
\renewcommand{\confidential}{}
\renewcommand{\anonsubtext}{(No author info supplied here, for consistency with
TACL-submission anonymization requirements)}
\newcommand{\instr}
\fi

\iftaclpubformat 

\else

\fi

\title{\rVQA: Diverse Visual Question Answering with Robustness Evaluation}

\author{Hannah Sterz$^1$~~~~~Jonas Pfeiffer$^2$~~~~~Ivan Vuli\'{c}$^1$ \\
$^1$Language Technology Lab, University of Cambridge\\
$^2$Google DeepMind 
}
  
\begin{document}
\maketitle
\begin{abstract}
Vision Language Models (VLMs) extend remarkable capabilities of text-only large language models and vision-only models, being able to learn from and process multi-modal vision-text input. While modern VLMs perform well on a number of standard image classification and image-text matching tasks, they still struggle with a number of crucial vision-language (VL) reasoning abilities such as counting and spatial reasoning. Moreover, while they might be very brittle to small variations in instructions and/or evaluation protocols, existing benchmarks fail to evaluate their robustness (or rather the lack of it). In order to couple challenging VL scenarios with comprehensive robustness evaluation, we introduce \textbf{\rVQA}, \textbf{D}iverse Visual Question \textbf{A}nswering with \textbf{R}obustness \textbf{E}valuation, a carefully created and curated multiple-choice VQA benchmark. \rVQA evaluates VLM performance on five diverse categories and includes four robustness-oriented evaluations based on the variations of: prompts, the subsets of answer options, the output format and the number of correct answers. Among a spectrum of other findings, we report that state-of-the-art VLMs still struggle with questions in most categories and are unable to consistently deliver their peak performance across the tested robustness evaluations. Consequently, our work calls for the systematic addition of robustness evaluations in future VLM research.

\end{abstract}

\section{Introduction and Motivation}
Building on the recent groundbreaking accomplishments of text-only Large Language Models (LLMs) across a wide range of (text-based) NLP tasks \cite{chiang2023vicuna,touvron2023llama}, there has been a growing interest in Vision-Language Models (VLMs) \cite{liu2023llava, zhu2023minigpt, team2023gemini, chen2023pali}. Such VLMs expand the input to comprise images as well as text, and enable dealing with cross-modal and multi-modal tasks and reasoning~\cite{flamingo,achiam2023gpt,liu2023llava}.




VLMs have been proven to perform well in standard image classification and image-text matching and reasoning tasks \cite{radford2021learning}. For instance, performance on the VQAv2 benchmark~\cite{balanced_vqa_v2}, which targets visual understanding and commonsense reasoning, is approaching human performance. The same also holds for benchmarks that require parsing and understanding of document images and text embedded into images such as TextVQA~\cite{singh2019towards} and DocVQA \cite{mathew2021docvqa}.  

However, there still exist particular vision-language (VL) reasoning problems which pose a challenge for the modern VLMs, such as counting and spatial reasoning \cite{onoe2024docci,li-etal-2024-topviewrs}. Furthermore, standard benchmarks fail to evaluate \textit{model robustness} to input and question variation,\footnote{In simple words, these benchmarks only include each question once and in a particular form, whereas a robust VLM should be able to answer correctly irrespective to different variations of the question where the correct answer can be clearly and unambiguously denoted (e.g., counting or determining spatial relations of objects).} a chosen inference protocol, as well as to specifications of the output and its format. High robustness to these aspects would ensure that the answer is not based on biases learned during training but due to a good understanding of the questions as well as of the instructions provided to the VLM. Put simply, we define robustness as the \textit{ability to perform consistently across such variations}.


To study and address these gaps, especially emphasising the robustness aspects, we carefully construct a new and challenging VQA benchmark, termed \textbf{D}iverse Visual Question \textbf{A}nswering with \textbf{R}obustness \textbf{E}valuation (\rVQA). It spans multiple-choice questions across five diverse, challenging scenarios/categories (e.g., conditional counting, visual commonsense, among others, see later in \S\ref{s:dataset}). \rVQA comes with the following key features. \textbf{1)} Diverse scenarios cover a range of crucial vision-language reasoning abilities needed by VLMs. \textbf{2)} Since current standard benchmarks are already largely saturated, they cannot (anymore) provide a good and insightful overview of the VLMs' abilities. Therefore, \rVQA comprises challenging and carefully curated evaluation instances from all five diverse scenarios. \textbf{3)} \rVQA provides the opportunity to study robustness within VQA evaluation.


\begin{figure*}[t!]
    \centering
    \includegraphics[width=0.91\linewidth]{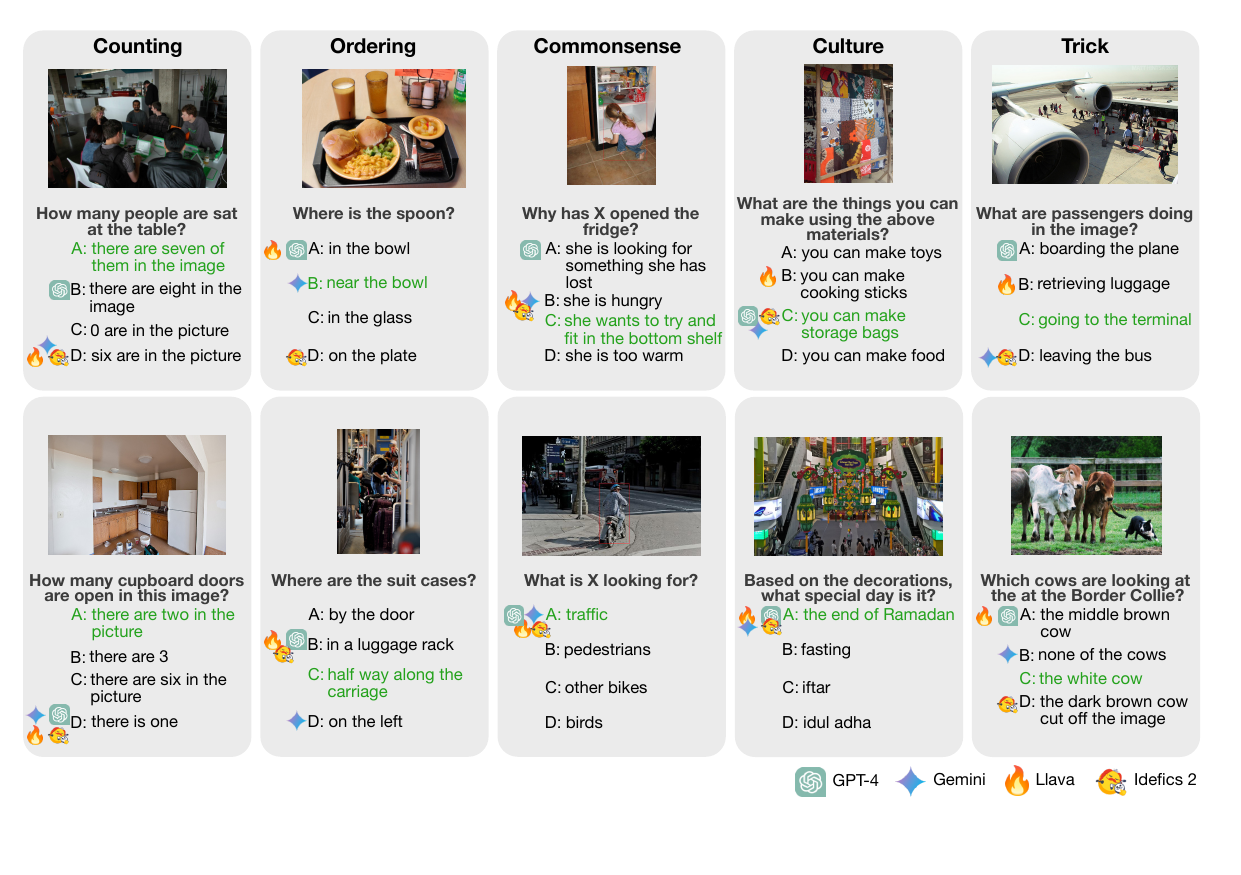}
    \vspace{-7mm}
    \caption{Examples of single-correct questions for each of the five categories covered in \rVQA.}
    \label{fig:examples}
    \vspace{-2mm}
\end{figure*}

We study robustness across multiple axes, that can roughly be grouped into \textbf{1)} variations of the prompt/instruction, \textbf{2)} variations of the set of answer options, \textbf{3)} variations of the output format, and \textbf{4)} variations of the number of correct answers; see later in \S\ref{ss:overview}. These axes of robustness, while typically neglected and discarded in current evaluations of VLMs, can unveil biases learned during pretraining and robustness to variations of the task, and potential inconsistencies of VLMs.

Unlike prior work on complex VL datasets where the primary purpose has been to evaluate complex visual understanding and language grounding \cite{lvis, winoground, tricd}, the scenarios covered by \rVQA do not rely on visually challenging images. Instead, they target reasoning abilities that are inherently easy for humans but require understanding of the question in the text form \textit{coupled} with image understanding; see Figure~\ref{fig:examples}. First, we include \textit{conditional counting} and \textit{spatial reasoning} as well-defined visual understanding tasks that VLMs are known to struggle with, and they require no additional (world) knowledge.\footnote{Conditional counting has the additional challenge that it requires multiple steps. First, all objects of a certain class need to be identified and filtered by some condition then the remaining objects need to be counted.} Two other scenarios require additional knowledge. \textit{Visual commonsense reasoning} questions require an understanding of the image and knowledge about the impact on the thoughts and actions of an individual. The \textit{culture} scenario requires knowledge about different cultures including their religions, dishes, and customs. The final, so-called \textit{trick} scenario is model-based and creates challenging questions by identifying mistakes in image descriptions created by most powerful, state-of-the-art VLMs, GPT-4 and Gemini. We believe that this combination of scenarios, while definitely non-exhaustive, covers a wide range of skills and can provide detailed insights into VLM performance when coupled with additional robustness challenges.


We then use \rVQA to examine and profile state-of-the-art closed-source and open-weights VLMs across the defined scenarios and robustness aspects, also aiming to trace their performance \textit{`chronologically'} by contrasting performance of most recent versus earlier checkpoints of the same VLM. As some main findings, we report that VLMs still struggle with the `simple-for-humans' vision understanding tasks of conditional counting and spatial reasoning, and they are not robust to variations in answer options. Even if the correct answer has the same content across all variations and is just phrased differently, the worst-case performance can be up to $33.6\%$ lower than the one observed with `vanilla' evaluation. All VLMs perform worse on questions with a varying number of possible correct answers. Especially LLaVA 1.6 and Idefics2 show a strong bias towards marking exactly one answer as correct. We also find that the generation and extraction strategies of the correct answers heavily impact performance: e.g., Gemini benefits from predicting the answers in JSON format, while LLaVA and Idefics2 do not perform well when prompted to provide JSON-formatted answers. 

 In hope to guide future developments of VLMs, we share \rVQA at \url{https://huggingface.co/datasets/cambridgeltl/DARE}.

\section{Background and Related Work}
\label{sec:rw}

%
%
\sparagraph{VQA: Preliminaries}
VQA is a task of providing the correct answer(s) given an image and a question about the image. In the standard \textit{multiple-choice} VQA variant, $n$ possible answer options to choose the correct answer(s) from are also provided, where one can differentiate between (a) the \textit{`single-correct'} setup (where it is known that there is always a single correct answer among the provided options) and (b) the \textit{`multi-correct'} setup\footnote{This differs substantially from the multiple-questions multiple-answers setup \cite{tang-etal-2024-multiple}, where multiple questions are processed at the same time and the model generates one answer for each question.} (or \textit{0-to-n} setup, where any number of correct answers from the provided options is allowed, including 0 and all $n$ of them). We follow this format and definitions of different setups in this work. 

\rparagraph{VQA Tasks and Datasets}
The VQA task was introduced in the VQA benchmark \cite{VQA}, where it consisted of short questions about the image that required visual understanding and commonsense. This dataset contains language bias,\footnote{Depending on the start of the question some answers are more likely: e.g., for a question starting with `what sport', `tennis' is the most likely answer, and correct in $41\%$ cases.} later addressed by VQAv2~\cite{balanced_vqa_v2}, which is still a standard benchmark despite the fact that it approaches human-level performance~\cite{chen2022pali}. 
%
%
GQA \cite{hudson2019gqa} provides questions about images covering targets addressing the semantic compositionality of scenes. This dataset is commonly treated via a classification task where all possible answers are possible classes. Consequently, this way the model can only answer questions where the correct answer is represented by one of the classes.

Several VQA datasets cover specific, finer-grained scenarios. VCR~\cite{zellers2019vcr} targets visual commonsense reasoning, which is the ability to make assumptions about plausible explanations for the motivation and next actions and thoughts. It uses movie scenes as images which results in a more limiting license of the dataset. CVQA \cite{romero2024cvqa} and CulturalVQA \cite{nayak-etal-2024-benchmarking} are VQA datasets that cover culturally diverse questions, limited to single-correct setups, and without any robustness analyses. TextVQA~\cite{singh2019towards}, DocVQA \cite{mathew2021docvqa} and InfographicVQA \cite{mathew2022info} target the ability to extract information of images of text, documents or infographics. VizWiz~\cite{gurari2018vizwiz} collects images and questions asked by blind people to interact with their environment. This includes object identification, colour detection or reading of texts. The model-based MMVP benchmark~\cite{tong_eyes_2024} targets weaknesses of current models by (i) identifying image pairs which have similar CLIP-computed embeddings \cite{radford2021learning} but different DINOv2-based embeddings~\cite{oquab2023dinov2}, and then (ii) annotating them with questions about the differences. 

Some benchmarks target multiple scenarios. The BLINK benchmark \cite{Fu2024Blink} focuses on visually challenging tasks such as finding bounding boxes for objects, image similarity, and camera movement between images. MMBench \cite{Liu2024MMBench} covers perception and reasoning questions. The majority of the samples focus on perception.

\vspace{0.4mm}
\noindent {\em Why \rVQA?}
Specialised datasets can only provide detailed insights into one single aspect of the VLMs performance. While benchmarks such as MMBench and BLINK target scenarios that are difficult for current VLMs, they focus on (direct) perception. With \rVQA we aim to cover scenarios that require reasoning such as filtering objects by a condition, spatial reasoning, and common sense reasoning. In the creation of \rVQA our goal is also to avoid the use of already established and (potentially and likely) saturated datasets, and to manually validate and curate all the included data instances to ensure high data quality.



\rparagraph{Robustness Evaluation}
Robustness refers to the ability to perform the same task across variations of the instances~\cite{jia-liang-2017-adversarial, dhole-etal-2023-nl, liang2022holistic, sclar2023quantifying}. These variations can be performed over a variety of aspects. In the scope of multiple-choice questions, \citet{wang2023large, pezeshkpour-hruschka-2024-large} show that the correct answer position within the options influences the performance of the model. Contrary to that, \citet{zheng2024large} find that not the order but rather the tokens indicating the options (e.g., A-D, 1-4) impact prediction. 




Another axis considered in some existing \textit{text-only} datasets is variations to the input. Variations such as deliberate typos, perturbations, and synonymy substitutions provide insights into the ability of the LLMs to respond to variations of the same question correctly~\cite{jia-liang-2017-adversarial, dhole-etal-2023-nl, liang2022holistic}. Such robustness tests then hint to worst-case model performance across the variations. While there are text-only language understanding benchmarks with such robustness measures vision-language benchmarks are just starting to include robustness.
MMBench \cite{Liu2024MMBench} performs a circular evaluation that prompts the model multiple times with the same question and shuffled options to reduce the impact of the answer position. However, this only covers a small set of possible variations of the VQA task.
To the best of our knowledge, \rVQA is the first vision-language benchmark to include robustness across several key axes as its crucial feature.





\section{\rVQA: Dataset Overview}
\label{s:dataset}
As introduced in \S\ref{sec:rw}, current VL benchmarks critically lack robustness evaluations: they only evaluate the task for specific instances. However, the VLM task performance may be heavily impacted by model biases and might thus not stem from actually understanding the provided text and images. We therefore create \rVQA, a multiple-choice VQA benchmark with multiple robustness evaluations. In a nutshell, the \rVQA dataset combines 5 different categories requiring reasoning and visual understanding, where all categories are framed as multiple-choice questions.\footnote{The multiple-choice format allows easy evaluation with accuracy metrics while still enabling coverage of a wide variety of questions with different topics and complexity.} We first briefly introduce the 5 categories (\S\ref{ss:categories}) followed by aspects of robustness evaluation (\S\ref{ss:overview}). A detailed description of \rVQA creation is provided in~\S\ref{ss:creation}. 



\subsection{Categories in \rVQA} 
\label{ss:categories}


The five categories in \rVQA cover a range of reasoning and vision understanding tasks. We include conditional counting and ordering to test vision understanding, one-hop reasoning and spatial reasoning. As categories that need (world) knowledge we include visual commonsense reasoning (VCR) and culture-based VQA. The VCR questions provide insights into the ability to understand the scene and infer the thoughts and motivations of a person in the scene. The culture category tests knowledge about different cultures. Finally, inspired by recent work~\cite{tong_eyes_2024}, the trick category targets shortcomings of the (currently) best-performing VLMs make in image descriptions. Some examples per each category are illustrated in Figure~\ref{fig:examples}. 

\rparagraph{Conditional Counting} 
These questions require counting objects based on an additional condition; it can be colour, position in the image, material, etc. Therefore, answering the questions requires not only identifying all objects of one type in the image, but also applying an additional filter on them before counting: e.g., the first example in Figure~\ref{fig:examples} requires identifying all people and then filtering only for people who are sitting.

\rparagraph{Ordering} It targets the ability to understand the order of objects and their spatial relation; e.g., the first example question in Figure~\ref{fig:examples} requires to identify the relation between the spoon and the other objects, whereas the second example shows a challenging data instance: one of the wrong answer options, \textit{`In a luggage rack'}, is where one would expect luggage a priori without seeing the image.

\rparagraph{Visual Commonsense Reasoning / VCR} In order to answer the questions, it is required to grasp the scene and make commonsense derivations from it about a person marked in the image, which might cover plausible motivation, thoughts, or the most likely next action of the person.

\rparagraph{Culture} Many objects and concepts encountered both in text and images are rooted in their culture and/or differ across different cultures~\cite{liu2024culturally}. This category thus requires answering questions about images of important concepts in different cultures, where we (incorrectly, for simplicity) approximate the culture by the language spoken by people. This requires the model to reason over concepts from a wide set of cultures. For instance, the examples in Figure~\ref{fig:examples} require knowledge about holidays, and use of certain objects.

\rparagraph{Trick} Here, we directly target mistakes in descriptions generated by state-of-the-art VLMs, where the key assumption is that the mistakes in the description point to incorrect understanding of that aspect of the image. To this end, we use GPT-4 and Gemini to obtain descriptions that we present to the annotators so that they can identify misconceptions and write questions targeting them. 
We note that through this design principle the questions are tied to the model used to generate the description. Therefore, this category should be used only for evaluation and comparison of models which were not used to generate the descriptions. The examples in this category cover a wide range of challenging questions (e.g., see examples in Figure~\ref{fig:examples}).

\begin{table}[t!]
    \centering
    \def\arraystretch{0.85}
   	\resizebox{1.0\linewidth}{!}{
    \begin{tabular}{lcccccc}
    \toprule
        & \multicolumn{3}{c}{\bf Validation} & \multicolumn{3}{c}{\bf Test}\\
        \cmidrule(lr){2-4} \cmidrule(lr){5-7}
         & \textit{Single} & \textit{Single+Var} & \textit{Multi} & \textit{Single} & \textit{Single+Var} & \textit{Multi}\\
         \cmidrule(lr){2-4} \cmidrule(lr){5-7}
         \textbf{Count} & 250 & 250 & 250& 250 & 250 & 250 \\
         \textbf{Order} & 231 & 228 & 250 & 237 & 237 &  250 \\
         \textbf{VCR} & 250 & - & -  & 258 & - & - \\
         \textbf{Culture} & - & - & - & 223 & 178 & 261   \\
         \textbf{Trick} & 232 & 232 & 250 & 229 & 229  &  250 \\
         \cmidrule(lr){2-4} \cmidrule(lr){5-7}
         \textbf{Total} & 963 & 710 & 750 & 1197 & 894 & 1011\\
        \bottomrule 
    \end{tabular}
    }
    \vspace{-1mm}
    \caption{The number of questions/samples across different evaluation scenarios (\textit{Single}: single correct answer, \textit{Single+Var}: single correct answer plus variations of the set of answer options (see \S\ref{ss:overview}), \textit{Multi}: multiple (0 to $n$) correct answers possible.}
    \label{tab:dataset_overview}
    \vspace{-1.5mm}
\end{table} 

\subsection{Overview of Robustness Evaluation Scenarios in \rVQA}
\label{ss:overview}

\sparagraph{Robustness} Task performance of LLMs and VLMs (as assessment of their abilities) may be substantially impacted by the prompt, examples, variations of the same text, and other factors~\cite{wei2022chain, liang2022holistic, lu-etal-2022-fantastically, zheng2024large}. Robustness is the ability to perform consistently across those variations. 

Variations in different aspects of the task require different skills from the model. Therefore, it is important to evaluate them in a variety of variants of the task to get a comprehensive understanding of model robustness. These variants can cover all aspect of the task. We include variants that are easy for a human as they do not change the content of the prompt or question. These variations cover the central parts of the task: prompt, answer options and output to determine the robustness to changes in these aspects. Moreover, we look at the robustness to the number of correct options which is a more challenging version of the task to evaluate the model's ability to adapt to changes in the structure of the task. Each of the four robustness axes requires different skills. This enables us to get a broad view on the model's robustness and what type of variations the model fails on.


\rparagraphnodot{Single-Correct Answer} is the standard, default scenario which serves as the reference point for all robustness evaluations. Each question has four answer options (by default marked with capital letters A-D) out of which exactly one is correct. The prompt provides the information that there is exactly one correct answer (see \ref{sec:prompts} for the full prompt). This setup is comparable to other multiple-choice datasets. The information allows one to pick the most likely answer without the need to decide the correctness of each option independently. {Starting from this basic setup, DARE currently includes four axes of variations that test model robustness:}

\rparagraph{(1) Varying the Sets of Answer Options} After collecting up to four correct and four incorrect options (see Appendix~\ref{ss:creation}), we sample three sets of four options where each set has a single correct option. The challenge in this setup is to answer the same question correctly more than once and to identify different correct answer options (among varying incorrect options as well). We report accuracy over all three variations. This provides the worst-case performance of the model for `unfortunate variations' of the data instances. Again, we provide the information that there is always a single correct answer as part of the prompt. We refer to this setup as \textit{Single-Correct Answer + Variations}. Table \ref{tab:dataset_overview} shows the number of samples for this setup (columns \textit{Single+Var}). 

\rparagraph{(2) Multiple Correct Answers \textit{(0-to-n)}} Here, we remove the requirement that there is only a single correct option compared to the \textit{Single-Correct Answer} setup. This requires the models to determine themselves the number of correct options: it is inherently a more challenging task as there are more options one can answer with, and there is fewer meta-information that can be used to rule out options. As a result, this is also more challenging for humans. 
For questions with multiple correct answer options, the performance on questions with one or more correct answers is different from the performance on questions with zero correct answers. Averaging the accuracy thus can lead to misleading results. For instance, a model that would simply label each question as not having an answer would get an accuracy $\sim20\%$. To avoid this conflation, we report results separately for question sets with $i=0,\ldots,4$ correct answers. 


\rparagraph{(3) Varying the Prompt}
 We use three different prompts in the \textit{Single-Correct Answer} and the \textit{Multiple-Correct Answer} scenarios (see~\ref{sec:prompts} for the prompts). We report the mean and standard deviation of the results on these prompts. 

The first prompt is a variation of the MMLU prompt~\cite{hendrycks2021measuring}, adapted to VQA multiple-choice questions. The second prompt introduces a scenario (\textit{`Imagine you are a student (...)'}) in which it is crucial to answer the questions correctly.\footnote{This prompt draws inspiration from the work of \citet{frohmann-etal-2024-segment}, which suggests and then empirically validates that using a prompt which defines a scenario that has high stakes for the model improves performance.} The third prompt describes the task factually, aiming to also minimise the lexical overlap with the first, default prompt.

\rparagraph{(4) Varying the Evaluation Protocol and Output Format} {In order to make the answers usable for other software for systems, the model must be able to provide them in a specified format. We test the ability of the model to provide the answer in structured formats such as JSON and CSV, and compare it against standard evaluation protocols for multiple-choice questions such as (i) `running text' generation and (ii) direct comparison of logits.} In the first, `vanilla' setup we prompt the model to provide the answer as a single character (i.e., choosing from A-D) or a list of characters, depending on the number of correct options expected (single-correct versus multi-correct setups). Given that VLMs are typically also not very robust to formatting even this simple output according to the provided instructions (i.e., we can encounter variants of the output such as \textit{B}, \textit{B.}, \textit{B:} etc.), we manually define a regular expression aimed to regularise this variation and extract the final, normalised answer (see Appendix~\ref{sec:regex}). This variant is termed \textsc{Out-Gen}. 


The second variant uses logit probabilities of output characters denoting the answer options. 
We limit the set of tokens the model can output to the tokens enumerating the options, that is, A, B, C, and D. For the single-correct setup, we let the model generate exactly one of these tokens. To extend this to the multi-correct setup, we include the \textit{end-of-generation} token. In this scenario, the output length is not limited to one. We let the model greedily generate tokens until the end-of-generation token obtains the highest probability or the maximum number of tokens is reached. Each option corresponding to generated tokens is labelled as correct. This variant, termed, \textsc{Out-Log}, requires access to logit probabilities, which is not available for many closed-source API-gated models.


We can also prompt the models to provide the answers in a valid JSON format (e.g., an example output is \textit{\{``answer'': ``B''\}}) - this variant is referred to as \textsc{Out-JSON}. As a side experiment, we further probe robustness by experimenting with a small selection of other possible output formats (including instructions which are very easy for humans to comprehend) later in \S\ref{sec:results}. The exact prompts for all the variants are listed in Appendix~\ref{sec:appendix}.

\rparagraph{Impact on Task Difficulty} Robustness evaluations change the circumstances of the task to test whether the model can maintain performance. These changes might have consequences on task complexity and difficulty. In the following, we discuss the impact of the introduced robustness variations on task difficulty. 


All used \textit{prompts} contain the same information; thus the prompt does not affect task difficulty.  The \textit{output format} needs to be structured or constrained so that a parsing algorithm can extract the options from the generated text. We choose the formats so they are easy to interchange for humans. For instance, the prompts for the JSON format include an example that enables someone without knowledge of JSON to provide the correct answer(s). However, these variations, even though they are easy for humans, could still result in changed task difficulty.

To create the \textit{sets of answer options}, the options are drawn randomly from the annotations. Therefore, we assume that the different subsets have the same difficulty on average. The \textit{varying number of correct options} variation increases difficulty notably. The chance of correctly guessing is $25\%$ for single-correct setups. However, by allowing multiple or no correct answers, the set of correct answers is the power set of {A, B, C, D} with a probability of a correct guess of $6.25\%$. 

\subsection{Data Creation and Annotation Process}
\label{ss:creation}
\sparagraph{Image Selection}
For the \textit{conditional counting}, \textit{ordering}, \textit{VCR}, and \textit{trick} categories, we manually select 500 images from the COCO 2017 dataset~\cite{lin2014microsoft}, with 250 images taken from its development set, and another 250 from the test set. The main (albeit subjective) selection criterion is that the images must show scenes that provide the grounds for challenging questions that are aligned with the corresponding categories.\footnote{For instance, close-ups of individual objects typically \textit{do not allow} for the construction of challenging questions that require counting, spatial reasoning or VCR.} 
As a necessary extra step in \textit{VCR}, we use DETR~\cite{carion2020end} to mark the subject of the question with a bounding box (denoted by X in the question, see Figure~\ref{fig:examples}). The boxes are manually validated to ensure that they mark a clearly visible person.
For \textit{trick}, we generate descriptions with Gemini~\cite{team2023gemini} and GPT-4 \cite{achiam2023gpt}, each model describing one half of the development set and one half of the test set.

The \textit{culture} category comprises images sampled from the MaRVL dataset~\cite{liu-etal-2021-visually}, following a similar procedure as with COCO. We approximate cultures by languages and select, again manually, a diverse subset of images for each of the five languages, representing a variety of concepts. We annotate the same number of images for each language. After quality control this results in 61 `Chinese', 38 `Swahili', 57 `Tamil', 60 `Turkish', and 45 `Indonesian' samples.
Since COCO and MaRVL images are available online, we cannot rule out that VLMs for which their training data are not disclosed have seen the images during pretraining. To prevent contamination, we do not publish the annotations of the annotated test set of \rVQA.

\rparagraph{Question Creation} 
We present human annotators\footnote{All annotators are hired through Prolific and compensated with $12\$$ per hour. We require annotators to have English as their first language and have a high-school diploma or higher education. The language requirement for cultural questions (during question creation, quality control) is adapted to being fluent in English and the language associated with the culture, to ensure annotators are familiar with the culture.} with an image and ask them to write a question for one of the categories, where each category comes with its own, customised set of annotation instructions with some examples (see Appendix~\S\ref{sec:annotation_guidelines} for details). For the conditional counting category, we ask them to provide the question and the correct number as the answer. This way, we can generate correct and incorrect answer options from templates to get a multiple-choice setup consistent with the other categories. For the ordering and trick categories, we ask the annotators to provide the question as well as four correct answers and four incorrect answers. We employ a similar setup for the culture category, but we require the annotators to be fluent speakers of the language the culture is approximated by. This ensures that they are familiar with the concepts shown in the image. The commonsense questions ask for the most plausible answer. We ask annotators to provide the question, a single, most plausible answer and three other plausible answers, and take the most plausible option as the single correct answer.\footnote{Since for VCR we consider the most plausible option as the correct answer, this category cannot be easily transferred into the scenario with multiple correct answers: i.e., the plausibility threshold that determines correct answers would be very subjective and we would expect humans to perform poorly on that task as well. Therefore, we skip the robustness test with multiple correct answers for VCR, cf. Table~\ref{tab:dataset_overview}.} 

\rparagraph{Quality Control}
To ensure high quality in \rVQA, we run a quality control stage, where we ask another set of annotators to validate the annotated samples. For \textit{conditional counting}, we ask two annotators to answer the questions with the correct number. If both agree with the (original) annotation we include it in the dataset and generate answer options using predefined templates to obtain a multiple-choice question. If there is a tie between the annotators we ask a third annotator to answer the question and include the data instance into \rVQA iff they also agree with the original correct answer and one of the two competing annotators. When dealing with categories which can have multiple correct (and incorrect) answers, we present each option separately with the question and the corresponding image and ask the annotator to decide whether the answer is correct or not for the given question and image. We get the answers from two annotators and include a third one if they disagree. We keep only questions where two control annotators agree with the original annotation.

For \textit{commonsense}, the notion of a correct answer is different than for the other categories: all answer options should be plausible. The answer that should be selected is then the most plausible one. We employ majority voting to determine which answer option is considered the most plausible by most annotators. This can be different from what the annotator intended as the most plausible option.

\rparagraph{Human Performance} We evaluate human performance on the DARE test set by sampling 100 samples for each category and assigning humans \footnote{These annotators are also hired through prolific with the same requirements as data creation and quality control.} to provide answers. As the multi-correct setup is more difficult than the single-correct setup, we asses the human baseline for both scenarios separately. 
\begin{table}[t!]
    \centering
    {\footnotesize
    \begin{tabularx}{0.99\linewidth}{lXXXXX}
        \toprule
         & Count & Order & VCR & Culture & Trick  \\
         \cmidrule(lr){2-6}
         Human Acc. & 92 & 96 & 70 & 82 & 92\\
         \bottomrule
    \end{tabularx}
    }%
    \vspace{-0.5mm}
    \caption{Human performance (accuracy) in the single-correct answer setup.}
    \label{tab:human_one_correct}
    \vspace{-1mm}
\end{table}
Human performance in the single-correct setup is shown in Table~\ref{tab:human_one_correct}. Humans seem to find categories \textit{conditional counting}, \textit{order}, and \textit{trick} easy, with corresponding accuracies $>90\%$. The lowest accuracy is achieved for the VCR category. Here, the correct answer is the most plausible out of a set of plausible options, which makes it more subjective and correspondingly more challenging than the other categories where the answers are much more clear-cut. 
\begin{table}[t!]
    \centering
    \def\arraystretch{0.85}
    {\footnotesize
    \begin{tabularx}{0.99\linewidth}{lXXXXXX}
        \toprule
         & 0 & 1 & 2 & 3 & 4 & F-1  \\
         \cmidrule(lr){2-6} \cmidrule(lr){7-7}
         Count & 77 & 86 & 81 & 89 & 79 & 89\\
         Order & 67 & 87 & 68 & 59 & 59 & 88\\
         Culture & 59 & 67 & 44 & 18 & 41 & 73 \\
         Trick & 82 & 74 & 54 & 57 & 41 & 85\\
         \bottomrule
    \end{tabularx}
    }%
    \vspace{-0.5mm}
    \caption{Human performance in the multi-correct setup; accuracy per question group (based on the number of correct answers) and averaged F-1 scores reported.}
    \label{tab:human_n_correct}
    \vspace{-1mm}
\end{table}

Human performance in the more challenging multi-correct setup is shown in Table~\ref{tab:human_n_correct}. As expected, absolute performance drops compared to the single-correct setup. Foreshadowing, human performance for  \textit{conditional counting}, \textit{order}, and \textit{trick} categories is clearly stronger than performances of the VLMs in our evaluation (cf., \S\ref{sec:results} later). For the \textit{culture} category, annotators show a tendency to only select one correct answer and not consider the other options, which is the main reason behind slightly lower scores for that category.

\rparagraph{Final Dataset}
{We note that we apply only the prompt variations and a subset of the output formats to the questions in the multi-correct scenario.\footnote{Varying the set of answer options is not possible while keeping the number of correct options fixed for each question. This would introduce multiple variations at once, which is why we focus only on variations in prompt and output format for multi-correct setups, and we look at the performance for the different number of correct options separately.}}

The final data statistics over different categories and setups are provided in Table~\ref{tab:dataset_overview}. Unless stated otherwise, we always report the results on the test portion, for which the correct answers are not publicly available to prevent data contamination~\cite{dong-etal-2024-generalization}. The validation portion, with the correct answers available, is intended for hyper-parameter tuning and local evaluation. Along with the final dataset, we also release the annotations from which we sample the setups from Table~\ref{tab:dataset_overview}; this might enable creation of new robustness evaluations by other researchers as well, beyond the ones introduced in our work.

\section{Experiments and Results}
\label{sec:results}

\begin{table*}[t!]
    \centering
    \def\arraystretch{0.95}
   	\resizebox{\linewidth}{!}{
    \begin{tabular}{lrrrrrrrrrrrrrrr}
    \toprule
         & \multicolumn{3}{c}{\bf Count} & \multicolumn{3}{c}{\bf Order} & \multicolumn{3}{c}{\bf VCR} & \multicolumn{3}{c}{\bf Culture} & \multicolumn{3}{c}{\bf Trick} \\
         \cmidrule(lr){2-4}\cmidrule(lr){5-7} \cmidrule(lr){8-10} \cmidrule(lr){11-13} \cmidrule(lr){14-16} 
         & \textsc{Gen} & \textsc{Log} & \textsc{json}  & \textsc{Gen} & \textsc{Log} & \textsc{json}  & \textsc{Gen} & \textsc{Log} & \textsc{json}  & \textsc{Gen} & \textsc{Log} & \textsc{json}  & \textsc{Gen} & \textsc{Log} & \textsc{json}  \\
          \cmidrule(lr){2-4}\cmidrule(lr){5-7} \cmidrule(lr){8-10} \cmidrule(lr){11-13} \cmidrule(lr){14-16} 
          \textbf{GPT-4} & $46.1_{\pm4}$ & $40.9_{\pm4}$ & $46.3_{\pm1}$ & $52.8_{\pm2}$ & $52.0_{\pm\mathbf{1}}$ & $51.9_{\pm2}$ & $\mathbf{63.1}_{\pm2}$ & $\mathbf{60.6}_{\pm2} $ & $63.2_{\pm1}$ & $78.3_{\pm6}$ & $83.4_{\pm2}$ & $86.4_{\pm1}$ & $53.7_{\pm1}$ & $55.7_{\pm2}$ & $55.0_{\pm2}$\\ 
          \textbf{Gemini} & $\mathbf{55.5}_{\pm\mathbf{2}}$ & - & $\mathbf{53.7}_{\pm9}$ & $\mathbf{71.6}_{\pm11}$  & - & $\mathbf{74.3}_{\pm6}$& $60.7_{\pm3}$ & - & $61.5_{\pm1}$ & $\mathbf{83.8}_{\pm7}$ & - & $\mathbf{87.9}_{\pm2}$& $\mathbf{70.0}_{\pm5}$ & - & $\mathbf{71.5}_{7}$\\
          \textbf{Llava 1.6} & $42.0_{\pm8}$ & $40.9_{\pm4}$ & $30.8_{\pm20}$ & $48.7_{\pm9}$ & $45.9_{\pm2}$ & $38.1_{\pm2}$ & $51.2_{\pm\mathbf{1}}$ & $51.2_{\pm4}$&$29.6_{\pm23}$ & $72.2_{\pm\mathbf{3}}$ & $75.4_{\pm\mathbf{0}}$ & $51.9_{\pm4}$ & $51.7_{\pm3}$ & $51.2_{\pm3}$ & $45.6_{\pm4}$ \\
          \textbf{Idefics2} & $46.8_{\pm2}$ & $\mathbf{44.1}_{\pm\mathbf{1}}$ & $0.0_{\pm\mathbf{0}}$ & $59.1_{\pm\mathbf{1}}$ & $\mathbf{59.4}_{\pm1}$ & $0.0_{\pm\mathbf{0}}$ & $\mathbf{60.9}_{\pm2}$& $60.3_{\pm\mathbf{2}}$ & $0.0_{\pm\mathbf{0}}$ & $78.8_{\pm9}$ & $\mathbf{84.2}_{\pm1}$ & $0.0_{\pm\mathbf{0}}$ & $59.5_{\pm\mathbf{1}}$ & $\mathbf{57.3}_{\pm\mathbf{1}}$ & $0.0_{\pm\mathbf{0}}$\\
         \bottomrule
    \end{tabular}
    }
    \vspace{-0.5mm}
    \caption{Average accuracy and standard deviation on the questions with a single correct answer for the five categories across three different instruction prompts, and across three different output formats (\textsc{Gen}, \textsc{Log}, \textsc{JSON}, see \S\ref{ss:overview}).}
    \label{tab:test_1_correct}
    \vspace{-1mm}
\end{table*}


\begin{table*}[t!]
    \centering
    \def\arraystretch{0.9}
   	\resizebox{\linewidth}{!}{%
    \begin{tabular}{lcccccccccccc}
    \toprule
         & \multicolumn{3}{c}{\bf Count} & \multicolumn{3}{c}{\bf Order} & \multicolumn{3}{c}{\bf Culture} &\multicolumn{3}{c}{\bf Trick} \\
                   \cmidrule(lr){2-4}\cmidrule(lr){5-7} \cmidrule(lr){8-10} \cmidrule(lr){11-13}
         & \textsc{Gen} & \textsc{Log} & \textsc{json} & \textsc{Gen} & \textsc{Log} & \textsc{json} & \textsc{Gen} & \textsc{Log} & \textsc{json} & \textsc{Gen} & \textsc{Log} & \textsc{json}\\
          \cmidrule(lr){2-4}\cmidrule(lr){5-7} \cmidrule(lr){8-10} \cmidrule(lr){11-13}
          \textbf{GPT-4} & $46.9_{-1.5}$ & $15.2_{-22.4}$& $18.4_{-29.2}$ & $26.0_{-22.4}$ & $23.2_{-30.0}$ & $31.6_{-21.6}$ & $67.3_{-15.7}$ & $66.7_{-16.3}$ & $83.7_{-4.2}$ & $34.1_{-20.9}$ & $38.4_{-18.3}$ & $41.5_{-14.8}$\\
          \textbf{Gemini} & $36.4_{-20.8}$ & - & $36.0_{-23.6}$& $53.2_{-24.9}$ & - & $60.7_{-17.8}$& $41.1_{-34.2}$ & - & $76.6_{-11.3}$ & $38.8_{-31.1}$ & - & $61.1_{-14.9}$\\
          \textbf{Llava 1.6} & $16.8_{-32.4}$ & $11.6_{-33.6}$ & $15.6_{-29.2}$ & $23.6_{-30.4}$ & $21.5_{-25.8}$ & $17.7_{-19.4}$& $63.5_{-7.8}$ & $67.4_{-8.3}$ & $37.6_{-18.5}$& $23.6_{-30.1}$ & $34.5_{-20.5}$ & $30.4_{-11.1}$\\
          \textbf{Idefics2} & $18.8_{-26.0}$ & $18.8_{-26.0}$ & $0.0_{-0.0}$ & $31.0_{-28.5}$ & $32.9_{-25.7}$ & $0.0_{-0.0}$ & $51.7_{-32.7}$ & $83.1_{-1.7}$ & $0.0_{-0.0}$ & $37.5_{-22.3}$ & $41.0_{-16.2}$ & $0.0_{-0.0}$ \\
         \bottomrule
    \end{tabular}
    }
    \vspace{-0.5mm}
    \caption{Accuracy when models answer all variations on sets of answer options for the same question in the single-correct answer setup (i.e., \textit{Single-Correct + Variations} from \S\ref{ss:overview}) correctly, indicating worst-case performance and by default being lower or equal to the scores in the single-correct setup without variations. All the scores are reported with the default prompt, without any prompt variation. We also report the difference to the main results in the basic single-correct setup as the subscript , again with the same prompt.}
    \label{tab:test_1_var}
    \vspace{-1.5mm}
\end{table*}

\sparagraph{Model Selection}
We select a representative combination of closed API-gated and open-sourced models, focusing on the models considered state-of-the-art and with the strongest performance on previous vision-language benchmarks. The choice has been further motivated by the aims of \textbf{1)} getting a diverse perspective on their performance and robustness on diverse scenarios of \rVQA, and also \textbf{2)} the ability to trace model performance and progress `historically' by assessing the most up-to-date checkpoints of the models as well as their earlier checkpoints. We choose the following models: Gemini Flash, GPT-4, LLaVA 1.6 and Idefics 2 (see Appendix~\ref{ss:hyperparam} for technical details and hyper-parameters).

\rparagraph{Evaluation Metrics}
Unless stated otherwise, we report performance as \textit{accuracy} scores, capturing the proportion of correctly answered questions. For the multi-correct setup, to also take partially correct answers into account, we additionally provide averaged F-1 scores: they are computed on the option level based on the recall and precision of identifying correct options.  

\subsection{Main Results and Discussion}
We now discuss the results across different categories and robustness aspects discussed in \S\ref{s:dataset}.

\rparagraph{Single-Correct Answer Setup} 
The results, summarised in Table~\ref{tab:test_1_correct}, indicate that even the simplest multiple-choice scenario is challenging for the state-of-the-art VLMs, with ample room for improvement. Gemini Flash shows the highest scores on average, followed by GPT-4 and Idefics2 which achieves similar results with \textsc{Out-Gen} and \textsc{Out-Log} as Gemini Flash and GPT-4. However, Idefics2 fails completely with \textsc{Out-JSON}. 

While the scores for all categories reveal substantial gaps to human-level performance except for \textit{culture}, \textit{conditional counting} seems especially challenging for all the VLMs in our evaluation, and across different evaluation protocols and formats. It is possible to attain higher absolute peak scores for the \textit{order} and \textit{trick} categories, but this is achieved with a subset of models coupled with specific evaluation protocols and output formats, suggesting some fundamental issues with robustness. %



\rparagraph{Robustness to Different Sets of Answer Options} We now test the proportion of questions for which the models can correctly answer all variations in the sets of answer options (i.e., the \textit{Single Correct Answer + Variations} setup). This provides an approximation of the worst-case model performance conditioned on the options provided to the model, with the results shown in Table~\ref{tab:test_1_var}.

Across all models, categories as well as output formats, performance in this setup is notably below the previous scores in the basic setup (cf., Table~\ref{tab:test_1_correct}). The gap is very substantial and this holds even for the \textit{conditional counting} category, where the correct answer is always the same number, which only gets expressed with different templates (e.g., \textit{3} versus \textit{there are 3}, see some examples in Figure~\ref{fig:examples} again). For the other scenarios, the different variations can also provide different correct answers (e.g., a position relative to different objects in an image for the \textit{Ordering} category), which also adds to task complexity. Absolute scores and gaps, as expected, also depend on the model and the chosen output format (e.g., Idefics2 has a smaller performance drop with \textsc{Out-Log} than with \textsc{Out-Gen}).

For the \textit{Culture} category, the performance gap is smaller and for specific configurations such as Idefics2 with \textsc{Out-Log} the gap is even less than $2\%$. However, this does not hold in general, but only for some models and configurations, e.g., Idefics2 with \textsc{Out-Gen} does show a substantial gap. This observation further highlights the importance of evaluating robustness along with performance, to ensure that VL understanding and reasoning capabilities are not limited only to specific scenarios, evaluation protocols, or task instances.

\begin{table*}[t!]
    \centering
    \def\arraystretch{0.83}
    \resizebox{\textwidth}{!}{%
    \begin{tabular}{llrrrrrrrrrrrrrrrrrr}
    \toprule
          &  & \multicolumn{3}{c}{\bf 0} & \multicolumn{3}{c}{\bf 1} & \multicolumn{3}{c}{\bf 2} & \multicolumn{3}{c}{\bf 3} & \multicolumn{3}{c}{\bf 4} &  \multicolumn{3}{c}{\bf F1} \\
          \cmidrule(lr){3-5}\cmidrule(lr){6-8} \cmidrule(lr){9-11} \cmidrule(lr){12-14} \cmidrule(lr){15-17} \cmidrule(lr){18-20}
          & & \textsc{Gen} & \textsc{Log} & \textsc{json} & \textsc{Gen} & \textsc{Log} & \textsc{json} & \textsc{Gen} & \textsc{Log} & \textsc{json} & \textsc{Gen} & \textsc{Log} & \textsc{json} & \textsc{Gen} & \textsc{Log} & \textsc{json} & \textsc{Gen} & \textsc{Log} & \textsc{json} \\
         \cmidrule(lr){3-5}\cmidrule(lr){6-8} \cmidrule(lr){9-11} \cmidrule(lr){12-14} \cmidrule(lr){15-17} \cmidrule(lr){18-20} 
          \multirow{4}{*}{\rotatebox[origin=l]{90}{\textbf{Count}}}& GPT 4 & $19.6$ & $35.5$ & $11.6$ & $41.9$ & $0.7$ & $38.1$ & $37.0$ & $16.1$ & $51.6$ & $33.1$ & $40.2$ & $54.2$ & $10.0$ & $0.0$ & $31.3$ & $58.4$ & $57.8$ & $62.3$\\
        & Gemini & $28.6$ & $-$ & $14.0$ & $48.5$ & $-$ & $51.6$ & $54.6$ & $-$ & $43.6$ & $51.9$ & $-$ & $86.0$ & $65.4$ & $-$ & $93.9$ & $74.9$ & $-$ & $82.1$\\
        & LLaVA 1.6 & $0.0$ & $0.0$ & $66.0$ & $29.3$ & $39.3$ & $9.3$ & $0.7$ & $0.0$ & $2.7$ & $0.0$ & $0.0$ & $0.0$ & $2.6$ & $25.0$ & $0.0$ & $41.8$ & $55.3$ & $14.6$\\
        & Idefics 2 & $0.0$ & $0.0$ & $100.0$ & $42.0$ & $40.7$ & $0.0$ & $0.0$ & $1.3$ & $0.0$ & $2.0$ & $6.8$ & $0.0$ & $4.6$ & $28.1$ & $0.0$ & $41.9$ & $60.2$ & $0.0$\\
         \midrule
          \multirow{4}{*}{\rotatebox[origin=l]{90}{\textbf{Order}}}& GPT 4 & $16.7$ & $22.7$ & $18.7$ & $33.3$ & $2.0$ & $36.7$ & $20.7$ & $4.0$ & $24.0$ & $5.3$ & $24.0$ & $8.7$ & $0.0$ & $0.7$ & $2.7$ & $55.3$ & $59.6$ & $58.0$\\
        & Gemini & $24.0$ & $-$ & $14.0$ & $50.0$ & $-$ & $56.0$ & $39.3$ & $-$ & $46.0$ & $10.0$ & $-$ & $25.3$ & $14.7$ & $-$ & $11.3$ & $66.1$ & $-$ & $72.8$\\
        & LLaVA 1.6 & $0.0$ & $0.0$ & $80.7$ & $30.7$ & $44.7$ & $9.3$ & $0.0$ & $0.0$ & $1.3$ & $0.0$ & $0.0$ & $0.0$ & $1.3$ & $18.7$ & $0.0$ & $38.8$ & $49.6$ & $13.9$\\
        & Idefics 2 & $0.0$ & $0.0$ & $100.0$ & $52.0$ & $40.7$ & $0.0$ & $2.7$ & $6.0$ & $0.0$ & $0.0$ & $7.3$ & $0.0$ & $0.0$ & $22.0$ & $0.0$ & $47.2$ & $56.4$ & $0.0$\\
         \midrule
         \multirow{4}{*}{\rotatebox[origin=l]{90}{\textbf{Culture}}}& GPT 4 & $54.2$ & $62.2$ & $35.8$ & $55.9$ & $21.4$ & $67.9$ & $42.7$ & $20.2$ & $54.9$ & $36.4$ & $53.3$ & $41.0$ & $19.0$ & $5.1$ & $17.7$ & $71.0$ & $71.5$ & $78.1$\\
        & Gemini & $56.9$ & $-$ & $20.8$ & $60.2$ & $-$ & $75.6$ & $43.5$ & $-$ & $52.9$ & $39.0$ & $-$ & $62.9$ & $28.4$ & $-$ & $53.4$ & $78.7$ & $-$ & $84.9$\\
        & LLaVA 1.6 & $0.0$ & $0.0$ & $77.1$ & $66.2$ & $73.1$ & $8.3$ & $0.6$ & $0.0$ & $5.7$ & $0.0$ & $14.8$ & $0.0$ & $15.1$ & $36.4$ & $0.0$ & $51.2$ & $61.4$ & $14.7$\\
        & Idefics 2 & $11.3$ & $2.7$ & $100.0$ & $72.5$ & $63.0$ & $0.0$ & $5.7$ & $14.4$ & $0.0$ & $4.5$ & $21.1$ & $0.0$ & $6.9$ & $30.7$ & $0.0$ & $51.1$ & $66.3$ & $0.0$\\
         \midrule
          \multirow{4}{*}{\rotatebox[origin=l]{90}{\textbf{Trick}}} & GPT 4 & $21.7$ & $30.7$ & $11.5$ & $38.0$ & $6.0$ & $44.7$ & $35.2$ & $13.2$ & $34.6$ & $12.7$ & $24.0$ & $24.7$ & $8.7$ & $1.3$ & $10.7$ & $63.1$ & $59.8$ & $61.1$\\
        & Gemini & $30.5$ & $-$ & $2.7$ & $46.0$ & $-$ & $56.0$ & $32.0$ & $-$ & $41.7$ & $20.0$ & $-$ & $36.7$ & $12.0$ & $-$ & $28.7$ & $64.3$ & $-$ & $72.1$\\
        & LLaVA 1.6 & $0.0$ & $0.0$ & $71.4$ & $44.0$ & $46.0$ & $10.0$ & $1.3$ & $0.0$ & $3.3$ & $0.0$ & $0.0$ & $0.0$ & $2.0$ & $26.0$ & $0.0$ & $42.2$ & $51.1$ & $14.6$\\
        & Idefics 2 & $0.7$ & $0.0$ & $100.0$ & $62.7$ & $48.7$ & $0.0$ & $4.6$ & $4.6$ & $0.0$ & $2.7$ & $8.7$ & $0.0$ & $2.0$ & $23.3$ & $0.0$ & $48.0$ & $58.8$ & $0.0$\\
         \bottomrule
    \end{tabular}
    }
    \vspace{-0.5mm}
    \caption{Accuracy scores in the Multi-Correct Setup, showing the proportion of questions for which the models detected all the answers correctly, split over the questions based on the number of gold correct answers (from 0 to 4); F1 scores are also reported to account for partially correct answers as well. We omit standard deviation over 3 prompts for clarity of presentation: for that, see Table~\ref{tab:test_n_correct_detail_prompts}.}
    \label{tab:test_n_correct_main}
    \vspace{-1.5mm}
\end{table*}

\rparagraph{Multi-Correct Answer Setup} 
This setup requires the model to reason separately over the answer options as the number of correct answer is not limited to a single option, and can also be zero. The results, split over groups of questions based on the number of correct answers expected, are provided Table~\ref{tab:test_n_correct_main}, with standard deviation due to varying the prompt provided in Table~\ref{tab:test_n_correct_detail_prompts} in Appendix~\ref{app:additional}.

First, we note that the results for the questions with one correct answer (i.e., comparable to the previous single-correct setup) are now lower in this, more challenging setup as the model lacks the prior information on the number of correct answers. 


\begin{figure}[t!]
    \centering
    \includegraphics[width=0.83\linewidth]{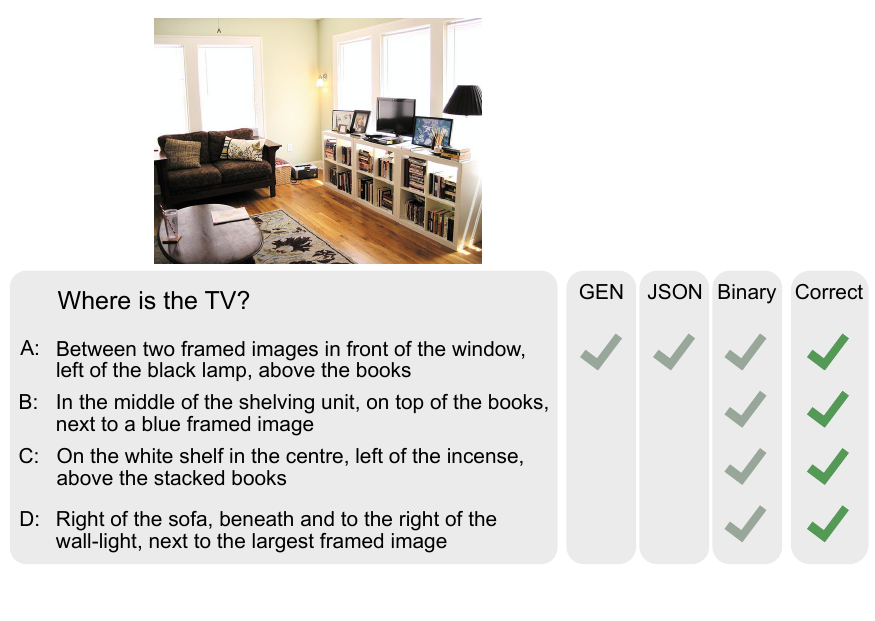}
    \vspace{-3mm}
    \caption{An example question from the \textit{order} category that Gemini 1.5 Flash answers incorrectly with the \textsc{Out-GEN} and the \textsc{Out-JSON} format with all options presented in a single prompt, but gets correctly when framed as a binary task for each individual option.}
    \label{fig:binary_example}
    \vspace{-3mm}
\end{figure}

Looking into performance of specific models, Gemini Flash seems to perform the best overall, followed by GPT-4. However, a closer inspection again reveals major problems with robustness, as peak absolute scores are achieved only with specific configurations and can vary considerably over different output formats and question types. For instance, \textsc{Out-Log} substantially outperforms \textsc{Out-Gen}. For \textsc{Out-Log}, the structure of the answer is integrated with the evaluation protocol, which might positively impact performance. Furthermore, for some models in some categories, \textsc{Out-JSON} yields the highest absolute scores (e.g., both Gemini Flash and GPT-4 display the highest F-1 scores with \textsc{Out-JSON}). We also observe that LLaVA 1.6 and Idefics2 struggle with questions that do not have exactly one correct answer. We hypothesise that this is due to their pretraining bias, where they are skewed towards single-correct setups. 

{We further explore different, alternative setups for the multi-correct questions, aiming to address some of the previously detected gaps in multi-correct setups. First, adding \textit{‘All’} and \textit{‘None of the Above’} directly as extra answer options can mostly help with questions that comprise 0 or 4 correct answers; however, the models remain brittle and inconsistent behavior over models and groups of questions has been observed; see Table~\ref{tab:n_correct_all_none} in the appendix. Next, we particularly note the results with the so-called \textit{binary decisions} setup: it relies on making `local' decisions per each option. Put simply, VLMs are prompted to determine whether the option is correct as a binary decision for each option individually without information about other options. The results of this variant are provided in Table~\ref{tab:test_n_correct_binary} in the appendix. While independently querying the VLM for each option increases inference costs, it also improves performance across the board, and can reduce the bias to pick one correct option (see Figure~\ref{fig:binary_example} for an example).

Importantly, even when the
gains with this binary-decisions setup are achieved, there is still
ample room for improvement with all the VLMs in
the multi-correct setup. On average, for \textit{counting},
\textit{ordering} or \textit{culture} less than a third of the questions is completely solved by the state-of-the-art VLMs.

\begin{figure*}[t!]
    \centering
    \includegraphics[width=\linewidth]{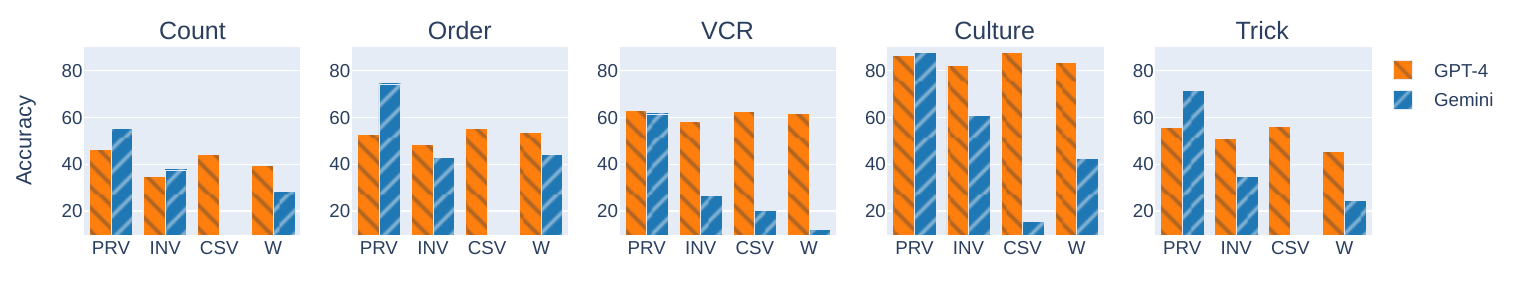}
    \vspace{-4.5mm}
    \caption{Evaluation across additional output formats (\textsc{Out}-\{\textsc{Inv}, \textsc{CSV}, \textsc{W=Word}\}, see \S\ref{ss:overview}) in the basic single-correct answer setup without variations. We do not report the results with LLaVA 1.6 and Idefics2 as they are straight zeros in all the evaluation runs for the three additional formats. The \textsc{Prv} column refers to the peak score for each model and category with any of the previously tested formats (\textsc{Out}-\{\textsc{Gen}, \textsc{Log}, \textsc{JSON}\} from Table~\ref{tab:test_1_correct}).}
    \label{fig:additional_output_formats}
    \vspace{-2mm}
\end{figure*}



\rparagraph{Robustness to Different Prompts} 
We measure this by the standard deviation of the scores across three prompts, both in single-correct (see Table~\ref{tab:test_1_correct} again) as well as in multi-correct setups (see Table~\ref{tab:test_n_correct_detail_prompts} in Appendix~\ref{app:additional}). In the single-correct setup, the differences in accuracy due to prompt variation range from $0$ to $20.4$, depending on other factors (i.e., the underlying model, category, output format). Gemini Flash, while reaching the highest average accuracy in most categories, at the same time tends to have a higher standard deviation than the other VLMs and is thus less robust to the chosen prompt. The highest standard deviation in this setup is with LLaVA 1.6 on \textit{conditional counting} with \textsc{Out-JSON}.  
%
Standard deviations seem even higher in the multi-correct setup (we relied on the `global' multiple-choice approach), again indicating that the scores are very volatile, and the VLMs seem even less robust to prompt variation here.

\rparagraph{Robustness to Output Formats} Further, based on the results from Tables~\ref{tab:test_1_correct}-\ref{tab:test_n_correct_main}, we observe that the performance of the VLMs varies substantially conditioned on the chosen evaluation protocol and the output format. As indicated before, Gemini Flash performs best with a JSON-formatted output (i.e., the \textsc{Out-JSON} variant), and it also suffers from inconsistent formatting with \textsc{Out-Gen}. Therefore, in order to investigate to what extent the detected lower performance is due to answer formatting not picked up by the manually defined post-processing regular expression (see \S\ref{ss:overview}),  we randomly sample 50 answers that are labeled as incorrect, and find that only 3/50 of the questions are false negatives. 

On the other hand, LLaVA's performance with \textsc{Out-JSON} drops notably compared to other evaluation protocols and formats, while Idefics2 is completely unable to provide JSON-formatted answers. This indicates a bias picked up during training of these VLMs. Namely, a general-purpose VLM with a true understanding of the prompt should be able to provide the answer in any format specified. However, LLaVA 1.6 and Idefics2 have been instruction-tuned to provide the letter indicating the correct answer option. This explains why they perform best with \textsc{Out-Gen} in most cases.

\vspace{0.3mm}
\noindent \textit{Probing Additional Output Formats.}
In order to further explore the robustness and flexibility of the VLMs to different output formats, we investigate three additional formats within the simplest, single-correct setup without variations: \textbf{1)} we `inversely' instruct the VLMs to generate a list of incorrect answers instead of a correct answer (labeled \textsc{Out-Inv}); \textbf{2)} similar to \textsc{Out-JSON}, we expect answers as valid CSV-formatted output (labeled \textsc{Out-CSV}); \textbf{3)} to further `stress-test' semantic understanding of the provided instructions, we propose a simple transformation where we expect the VLM to output any word from the English vocabulary that starts with the letter of the correct answer (\textsc{Out-Word}). The results with these additional output formats are summarised in Figure~\ref{fig:additional_output_formats}.

LLaVA 1.6 and Idefics2 are again completely unable to follow these output formats, which aligns with the previous observation concerning the JSON-formatted output. Gemini Flash struggles with the CSV format, whereas GPT-4 reaches peak scores with that format for 3/5 categories. We further observe substantial drops with Gemini Flash with the other two formats, while GPT-4 seems most robust when faced with the these three output formats, but again with drops compared to the previous three formats (the \textsc{Prv} column) from Table~\ref{tab:test_1_correct}, except for CSV. In sum, these results further point to general robustness and instruction-following issues with all the VLMs in our evaluation and again emphasise the gap to expected human-level performance.

\begin{figure*}[t!]
    \centering
    \includegraphics[width=0.94\linewidth]{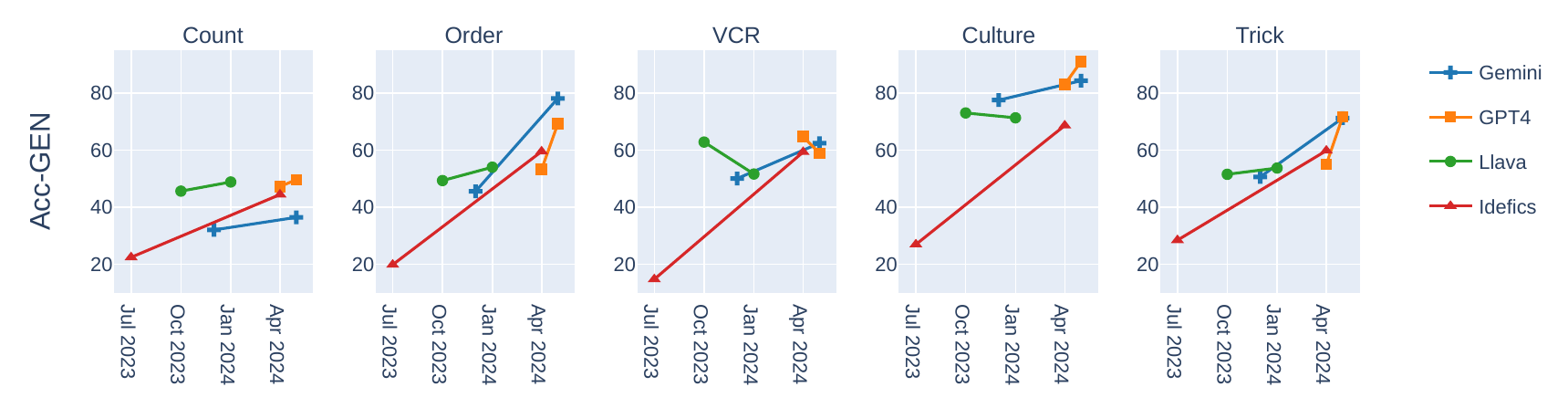}
    \vspace{-2mm}
    \caption{Tracing progress on the five categories of \rVQA. We assess (i) Gemini 1 and Gemini Flash 1.5; (ii) GPT-4 and GPT-4o, (iii) LLaVA 1.5 and LLaVA 1.6; (iv) Idefics1 and Idefics2.}
    \label{fig:development}
    \vspace{-2.5mm}
\end{figure*}

\subsection{Further Discussion and Analyses}
\label{ss:further}
As one general finding, we observe that different models perform well on different categories and in different configurations. Whereas Idefics2 shows less variation across different prompts, it struggles on the other three robustness evaluations. GPT-4 seems as the most robust model when varying the output format, while Gemini Flash outperforms other models in general in the multi-correct setup. Put simply, there is no general-purpose or `one-size-fits-all' solution, and all the models suffer from issues with robustness and flexibility, questioning their general semantic understanding.

Furthermore, we observe a gap between the API-gated models GPT-4 and Gemini Flash and the two open models. The former can (at least to some extent) handle different output formats and the multiple-correct setup substantially better than the latter. This hints that current open-weights models are still limited by tasks and output formats they observe during pretraining.

\rparagraph{Tracing Progress of VLMs}
To assess whether performance on the challenging questions from \rVQA improves over time and over newer VLM checkpoints, which would hint that the VLMs are indeed gradually improving, we evaluate different checkpoints of the same model in the single-correct answer setup (relying on the \textsc{Out-Gen} format, similar trends have been observed with the other output formats). The results are plotted in Figure~\ref{fig:development}, with further details on the evaluated checkpoints available in Appendix~\ref{sec:appendix_progress}.
%
There are some trends visible across all categories. Every model family shows improvements in the majority of categories, with only small or anecdotal drops (e.g., GPT-4 in VCR, LLaVA in the \textit{culture} category). LLaVA displays the smallest improvement while Idefics shows the steepest improvement; however, Idefics also starts from the lowest initial results. Gemini Flash 1.5 and GPT-4o show salient gains for the \textit{spatial ordering} and \textit{conditional counting} questions over their predecessors Gemini 1 and GPT4.


These results indicate that \rVQA with its coverage of different categories, evaluation setups and challenging scenarios could be used for benchmarking progress (or the lack of it as in the case of LLaVA) of different VLM families in the future: (i) while we detect gains with the newest checkpoints over the previous releases, (ii) there is still a large gap to human-level performance across the board, along with the wide array of robustness issues, as discussed in this paper. 





\section{Conclusion and Outlook}

We have introduced \rVQA, a novel VQA benchmark for vision-language models (VLMs) which targets image understanding and multi-modal reasoning across five diverse scenarios, offering challenging multiple-choice questions, which were carefully selected and manually annotated, and support evaluations in single-correct and multi-correct answer setups. \rVQA puts a special focus on \textit{evaluating various robustness aspects of the VLMs}, and it includes variations of the questions and evaluation protocols across several crucial axes (e.g., input prompt, output format, answer options).

Our extensive experiments on \rVQA highlight that modern VLMs still struggle with the robustness aspects, and their performance varies wildly and depends on multiple factors such as the evaluation setup (single-correct vs multi-correct), category complexity, evaluation protocol, specified output format, etc. VLMs that are constructed from visual encoders aligned to text-only LLMs with visual instruction tuning struggle the most in our robustness evaluations. Our work highlights the importance of including datasets that target task setups that are not directly covered in pretraining into their construction. In general, motivated by our experiments on tracing progress of VLMs over time (see Figure~\ref{fig:development} again), we hope that \rVQA will help guide future developments of VLMs, with a particular focus on increasing their flexibility to different variations covered in \rVQA and consequently mitigating their critical issues with robustness. 
\section*{Acknowledgments}
 Hannah Sterz thanks the Cambridge Trust for their support via the International Scholarship. This work has been supported by a Royal Society University Research Fellowship \textit{‘Inclusive and Sustainable Language Technology for a Truly Multilingual World’} (no 221137) awarded to Ivan Vuli\'{c}.
 We thank Aishwarya Kamath and Jeremiah Harmsen for thoughtful comments on initial drafts of the paper.



\bibliography{anthology,custom}
\bibliographystyle{acl_natbib}

\appendix

\section{Evaluation: Technical Details and Reproducibility}
\label{sec:appendix}
\subsection{Prompts}
\label{sec:prompts}
VLMs results heavily depend on the used prompt: for reproducibility, we report all prompts used during evaluation. For the \textbf{single-correct answer} setup we use the following prompt for \textsc{Out-GEN} and \textsc{Out-LOG}:

{\footnotesize
\begin{tcolorbox}[colback=gray!10, colframe=black, width=\linewidth, boxsep=4pt, left=2pt, right=2pt, top=2pt, bottom=2pt, title=Main prompt for multiple-choice questions with a single correct answer (\textsc{Out-GEN} and \textsc{Out-LOG})]
        The following are multiple choice questions about <QUESTION TYPE>. You should directly answer the question by choosing the correct option given the image and the question. Give only the letter indicating the correct answer e.g. "A"\\

        Question: <QUESTION> \\
        Options: \\
        A. <ANSWER A>\\
        B. <ANSWER B> \\
        C. <ANSWER C> \\
        D. <ANSWER D>  \\
        Answer:
\end{tcolorbox}
}%
{\footnotesize
\begin{tcolorbox}[colback=gray!10, colframe=black, width=\linewidth, boxsep=4pt, left=2pt, right=2pt, top=2pt, bottom=2pt, title=Prompt 2 for multiple-choice questions with a single correct answer (\textsc{Out-GEN} and \textsc{Out-LOG})]
        Imagine you are a student in an exam consisting of the following multiple choice question about <QUESTION TYPE>. [You need to do well in this exam in order to not fail the class.] Provide the answer by choosing the correct option, e.g. 'B'. \\

        Question: <QUESTION> \\
        Options: \\
        A. <ANSWER A>\\
        B. <ANSWER B> \\
        C. <ANSWER C> \\
        D. <ANSWER D>  \\
        Answer:
\end{tcolorbox}
}%
{\footnotesize
\begin{tcolorbox}[colback=gray!10, colframe=black, width=\linewidth, boxsep=4pt, left=2pt, right=2pt, top=2pt, bottom=2pt, title=Prompt 3 for multiple-choice questions with a single correct answer (\textsc{Out-GEN} and \textsc{Out-LOG})]
        You will be presented with a multiple choice question about <QUESTION TYPE>. Please return the answer as the letter corresponding to the correct option e.g. 'A'. \\

        Question: <QUESTION> \\
        Options: \\
        A. <ANSWER A>\\
        B. <ANSWER B> \\
        C. <ANSWER C> \\
        D. <ANSWER D>  \\
        Answer:
\end{tcolorbox}
}%

To get the VLM to return JSON-formatted answers we need to specify that requirement in the prompt.  We use the following prompts:

{\footnotesize
\begin{tcolorbox}[colback=gray!10, colframe=black, width=\linewidth, boxsep=4pt, left=2pt, right=2pt, top=2pt, bottom=2pt, title=Prompt 1 for multiple-choice questions with a single correct answer and output in json-format (\textsc{Out-JSON})]
        The following are multiple choice questions about <QUESTION TYPE>. You should directly answer the question by choosing the correct option given the image and the question. Provide the answer in json format e.g. \{"answer": "A"\}.\\

        Question: <QUESTION> \\
        Options: \\
        A. <ANSWER A>\\
        B. <ANSWER B> \\
        C. <ANSWER C> \\
        D. <ANSWER D>  \\
        Answer:
\end{tcolorbox}
}%
{\footnotesize
\begin{tcolorbox}[colback=gray!10, colframe=black, width=\linewidth, boxsep=4pt, left=2pt, right=2pt, top=2pt, bottom=2pt, title=Prompt 2 for multiple-choice questions with a single correct answer and output in json-format (\textsc{Out-JSON})]
        Imagine you are a student in an exam consisting of the following multiple choice question about <QUESTION TYPE>. [You need to do well in this exam in order to not fail the class.] Provide the answer by choosing the correct option as json e.g. \{"answer": "B"\}.\\

        Question: <QUESTION> \\
        Options: \\
        A. <ANSWER A>\\
        B. <ANSWER B> \\
        C. <ANSWER C> \\
        D. <ANSWER D>  \\
        Answer:
\end{tcolorbox}
}%
{\footnotesize
\begin{tcolorbox}[colback=gray!10, colframe=black, width=\linewidth, boxsep=4pt, left=2pt, right=2pt, top=2pt, bottom=2pt, title=Prompt 3 for multiple-choice questions with a single correct answer and output in json-format (\textsc{Out-JSON})]
        You will be presented with a multiple choice question about <QUESTION TYPE>. Please return the answer as json for instance \{"answer": "D"\}.\

        Question: <QUESTION> \\
        Options: \\
        A. <ANSWER A>\\
        B. <ANSWER B> \\
        C. <ANSWER C> \\
        D. <ANSWER D>  \\
        Answer:
\end{tcolorbox}
}%

For the \textbf{multi-correct} setup the prompt needs to specify that there can be a varying number of correct answer options.  We use the following prompts:

{\footnotesize
\begin{tcolorbox}[colback=gray!10, colframe=black, width=\linewidth, boxsep=4pt, left=2pt, right=2pt, top=2pt, bottom=2pt, title=Prompt 1 for multiple-choice questions with multiple correct answers (\textsc{Out-GEN} and \textsc{Out-LOG})]
        The following are multiple choice questions about <QUESTION TYPE>. You should directly answer the question by choosing the correct options given the image and the question. 
        There can be zero to four correct answers. If no answer is correct answer NONE otherwise provide a list of the correct answer options.\\

        Question: <QUESTION> \\
        Options: \\
        A. <ANSWER A>\\
        B. <ANSWER B> \\
        C. <ANSWER C> \\
        D. <ANSWER D>  \\
        Answers:
\end{tcolorbox}
}%
{\footnotesize
\begin{tcolorbox}[colback=gray!10, colframe=black, width=\linewidth, boxsep=4pt, left=2pt, right=2pt, top=2pt, bottom=2pt, title=Prompt 2 for multiple-choice questions with a multiple correct answer (\textsc{Out-GEN} and \textsc{Out-LOG})]
        Imagine you are a student in an exam consisting of the following multiple choice question about <QUESTION TYPE>. [You need to do well in this exam in order to not fail the class.] Provide the answer by choosing the correct option.
        Your teacher decided to make this exam extra hard. There can be zero to four correct answers. If there is no correct answer, answer with NONE otherwise return the list of correct options e.g. "A, C".\\

        Question: <QUESTION> \\
        Options: \\
        A. <ANSWER A>\\
        B. <ANSWER B> \\
        C. <ANSWER C> \\
        D. <ANSWER D>  \\
        Answers:
\end{tcolorbox}
}
{\footnotesize
\begin{tcolorbox}[colback=gray!10, colframe=black, width=\linewidth, boxsep=4pt, left=2pt, right=2pt, top=2pt, bottom=2pt, title=Prompt 3 for multiple-choice questions with a multiple correct answer (\textsc{Out-GEN} and \textsc{Out-LOG})]
        You will be presented with a multiple choice question about <QUESTION TYPE. Provide the correct answer given the question and the image as the number corresponding to the answer.
        Attention: There can be an arbitrary number of correct options. Return all that you identify as correct, e.g. "B, D". If no option is correct answer with NONE.\\

        Question: <QUESTION> \\
        Options: \\
        A. <ANSWER A>\\
        B. <ANSWER B> \\
        C. <ANSWER C> \\
        D. <ANSWER D>  \\
        Answers:
\end{tcolorbox}
}

The JSON-formatted output again requires a slightly different prompt:

{\footnotesize
\begin{tcolorbox}[colback=gray!10, colframe=black, width=\linewidth, boxsep=4pt, left=2pt, right=2pt, top=2pt, bottom=2pt, title=Prompt 1 for multiple-choice questions with a multiple correct answer with output in json-format (\textsc{Out-JSON})]
        The following are multiple choice questions about <QUESTION TYPE>. You should directly answer the question by choosing the correct options given the image and the question. 
        There can be zero to four correct answers. Provide the answer in json format e.g. \{"answers": ["A", "B"]\} or if there is no correct answer \{"answers": []\}.\\

        Question: <QUESTION> \\
        Options: \\
        A. <ANSWER A>\\
        B. <ANSWER B> \\
        C. <ANSWER C> \\
        D. <ANSWER D>  \\
        Answers:
\end{tcolorbox}
}
{\footnotesize
\begin{tcolorbox}[colback=gray!10, colframe=black, width=\linewidth, boxsep=4pt, left=2pt, right=2pt, top=2pt, bottom=2pt, title=Prompt 2 for multiple-choice questions with a multiple correct answer with output in json-format (\textsc{Out-JSON})]
        Imagine you are a student in an exam consisting of the following multiple choice question about <QUESTION TYPE>. [You need to do well in this exam in order to not fail the class.] Provide the answer by choosing the correct option.
        Your teacher decided to make this exam extra hard. There can be zero to four correct answers. There can be zero to four correct answers. Provide the answer by choosing the correct option as json e.g. \{"answer": ["B", "D"]\}.\\

        Question: <QUESTION> \\
        Options: \\
        A. <ANSWER A>\\
        B. <ANSWER B> \\
        C. <ANSWER C> \\
        D. <ANSWER D>  \\
        Answers:
\end{tcolorbox}
}
{\footnotesize
\begin{tcolorbox}[colback=gray!10, colframe=black, width=\linewidth, boxsep=4pt, left=2pt, right=2pt, top=2pt, bottom=2pt, title=Prompt 3 for multiple-choice questions with a multiple correct answer with output in JSON-format (\textsc{Out-JSON})]
        You will be presented with a multiple choice question about <QUESTION TYPE>. 0-4 answers can be correct. Please return the answer as json, for instance \{"answer": ["A", "D"]\}.\\

        Question: <QUESTION> \\
        Options: \\
        A. <ANSWER A>\\
        B. <ANSWER B> \\
        C. <ANSWER C> \\
        D. <ANSWER D>  \\
        Answers:
\end{tcolorbox}
}

{\footnotesize
\begin{tcolorbox}[colback=gray!10, colframe=black, width=\linewidth, boxsep=4pt, left=2pt, right=2pt, top=2pt, bottom=2pt, title=Prompt for multiple-choice questions with a single correct answer with output in inverse-format (\textsc{Out-INV})]
        The following is a question about <QUESTION TYPE>. Provide only the three incorrect answers to the question as the list of the corresponding letters e.g. 'A, C, D'.\\

        Question: <QUESTION> \\
        Options: \\
        A. <ANSWER A>\\
        B. <ANSWER B> \\
        C. <ANSWER C> \\
        D. <ANSWER D>  \\
        Answers:
\end{tcolorbox}
}
{\footnotesize
\begin{tcolorbox}[colback=gray!10, colframe=black, width=\linewidth, boxsep=4pt, left=2pt, right=2pt, top=2pt, bottom=2pt, title=Prompt for multiple-choice questions with a single correct answer with output in csv-format (\textsc{Out-CSV})]
        The following are multiple choice questions about <QUESTION TYPE>. You should directly answer the question by choosing the correct option given the image and the question. Provide the answer in csv format e.g. 'answer,A'.\\

        Question: <QUESTION> \\
        Options: \\
        A. <ANSWER A>\\
        B. <ANSWER B> \\
        C. <ANSWER C> \\
        D. <ANSWER D>  \\
        Answers:
\end{tcolorbox}
}
{\footnotesize
\begin{tcolorbox}[colback=gray!10, colframe=black, width=\linewidth, boxsep=4pt, left=2pt, right=2pt, top=2pt, bottom=2pt, title=Prompt 3 for multiple-choice questions with a single correct answer with output in word-format (\textsc{Out-WORD})]
       The following are multiple choice questions about <QUESTION TYPE>. You should answer the question by giving the answer as a word that starts with the letter of corresponding to the correct option e.g. if A is correct 'Apple'. Your answer has to start with that word.\\

        Question: <QUESTION> \\
        Options: \\
        A. <ANSWER A>\\
        B. <ANSWER B> \\
        C. <ANSWER C> \\
        D. <ANSWER D>  \\
        Answer:
\end{tcolorbox}
}

\subsection{Regex for Extracting Answers}
\label{sec:regex}
To extract the correct answer from the text we use regular expressions. The generated text is parsed for:
\verb/(^|\s)[ABCD](, [ABCD])+($|\n|,|\.|\s)/
to find a list of letters. To find single-letter answers the expression is parsed for:
\verb/(<OPT>: )|(<OPT>\.)|(<OPT>\n)|(<OPT>$)/ for all answer options: A, B, C and D, where \verb/<OPT>/ refers to the four options.

\subsection{Models \& Hyper-Parameters}
\label{ss:hyperparam}
\vspace{0.6mm}
\noindent \textbf{GPT-4} 
\cite{achiam2023gpt}, the API-gated model by OpenAI, supports text as well as image input. We use the \emph{gpt-4-turbo-2024-04-09} variant, unless stated otherwise. 

\vspace{0.4mm}
\noindent \textbf{Gemini} \cite{team2023gemini} is another API-gated model, created by Google, which supports textual and visual input. We use \emph{gemini-1.5-flash}, a lightweight variant targeting speed and efficiency.


\vspace{0.4mm}
\noindent \textbf{Idefics2} is an open-weights model that aligns OpenCLIP \citep{radford2021learning} with LLama \citep{touvron2023llama} with a Vision Language Connector. We use \emph{HuggingFaceM4/idefics2-8b}. 

\vspace{0.4mm}
\noindent \textbf{LLaVA} \cite{liu2023improvedllava} is an open-weights model that aligns the CLIP ViT-L/14 \citep{radford2021learning} embeddings with the Vicuna LLM \citep{chiang2023vicuna} via a projection of the image embeddings. This allows the model to handle text and image input jointly. We use \emph{llava-hf/llava-v1.6-vicuna-7b-hf}. 

For all the models, we opt for their default, suggested hyper-parameters: e.g., temperature is set to $0$ with GPT-4 and Gemini, while we use the default generation configuration of Idefics2 and LLaVa, corresponding to greedy decoding.

\subsection{Model Progress}
\label{sec:appendix_progress}
With frequent new versions of each model family, an interesting question is whether new versions already improve the score on our benchmark. To investigate this, we report the score of a previous or more recent model in the same family for the \textit{conditional counting} and \textit{order} categories. The development is illustrated in Figure \ref{fig:development}. We use \emph{gemini-1.0-pro-001} for Gemini 1.0, \emph{llava-hf/llava-1.5-7b-hf} for LLaVA 1.5, \emph{HuggingFaceM4/idefics-9b} for Idefics 1 and \emph{gpt-4o-2024-05-13} for GPT-4o. For the date of the model, we use the date of the first commit of the model to HuggingFace for models available on the HuggingFace hub. For API-gated models we use the dates specified in their documentation.

\begin{figure}[h!]
    \centering
    \begin{subfigure}[t]{0.5\textwidth}
        \centering
        \includegraphics[trim={3cm 0 3cm 0}, scale=0.3,page=1]{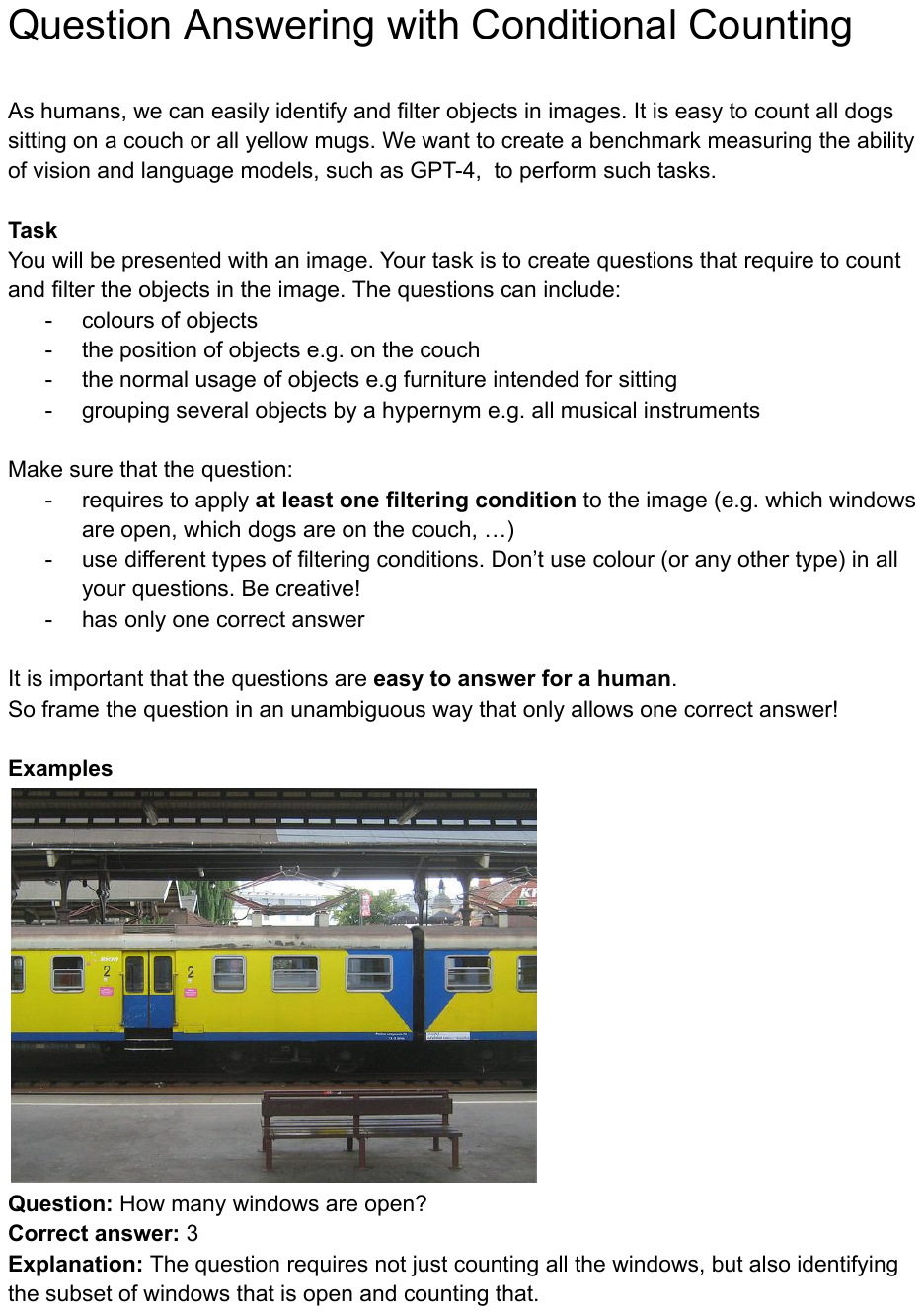}
    \end{subfigure}
    \begin{subfigure}[t]{0.5\textwidth}
        \centering
        \includegraphics[trim={3cm 17cm 3cm 4cm},scale=0.3,page=2]{images/appendix/guidelines_count.pdf}
    \end{subfigure}
    \caption{Guidelines for the \textit{conditional counting} category}
    \label{fig:annotation_count_1}
\end{figure}

\section{Annotation Guidelines}
\label{sec:annotation_guidelines}
We use guidelines customised for each category in \rVQA, with specific instructions and examples, to ensure that annotators pay attention to the most important aspects associated with each category. To illustrate the structure of the guidelines we report the guidelines for two of the categories. Figure \ref{fig:annotation_count_1} shows our guidelines for \textit{conditional counting}.

\section{Additional Results}
\label{app:additional}

\begin{table}[h!]
    \centering
    \def\arraystretch{0.9}
    \resizebox{0.9\linewidth}{!}{%
    \begin{tabular}{llrrrrrr}
    \toprule
          &  & \textbf{0} & \textbf{1} & \textbf{2} & \textbf{3} & \textbf{4} & \textbf{F1} \\
         \cmidrule(lr){3-7}  \cmidrule(lr){8-8}  
          \multirow{4}{*}{\rotatebox[origin=l]{90}{\textbf{Counting}}}& GPT 4  & $\textbf{86}_{+50}$ & $16_{-26}$ & $8_{-44}$ & $16_{-38}$ & $20_{-11}$ & $37.9_{-24}$ \\
          & Gemini & $13_{-11}$ & $\textbf{18}_{-38}$ & $14_{-32}$ & $\textbf{50}_{+25}$ & $65_{+51}$ & $69.3_{-3}$ \\
          & LLaVA 1.6 & $0_{-66}$ & $0_{-39}$ & $6_{+3}$ & $24_{+24}$ & $\textbf{92}_{+67}$ & $\textbf{69.6}_{+15}$ \\
          & Idefics2 & $32_{-68}$ & $16_{-26}$ & $\textbf{20}_{+19}$ & $24_{+17}$ & $44_{+16}$ &  $64.4_{+4}$ \\
         \midrule
          \multirow{4}{*}{\rotatebox[origin=l]{90}{\textbf{Order}}}& GPT 4  & $\textbf{72}_{-49}$ & $28_{-11}$ & $16_{-8}$ & $12_{+5}$ & $12_{+9}$ & $56.2_{-3}$\\
          & Gemini & $44_{+20}$ & $\textbf{40}_{-16}$ & $\textbf{54}_{+8}$ & $\textbf{50}_{+25}$ & $44_{+29}$ & $\textbf{80.2}_{+7}$\\
          & LLaVA 1.6 & $20_{-61}$ & $18_{-27}$ & $26_{+25}$ & $32_{+32}$ & $\textbf{60}_{+41}$ & $72.1_{+23}$\\
          & Idefics2 &  $24_{-76}$ & $10_{-42}$ & $22_{+16}$ & $32_{+25}$ & $42_{+20}$ & $69.1_{+13}$\\
         \midrule
          \multirow{4}{*}{\rotatebox[origin=l]{90}{\textbf{Culture}}} & GPT 4  &  $\textbf{94}_{+32}$ & $58_{-10}$ & $36_{-10}$ & $29_{-24}$ & $32_{+15}$ & $\textbf{83.0}_{+5}$\\
          & Gemini & $84_{+27}$ & $\textbf{63}_{-13}$ & $\textbf{55}_{+2}$ & $\textbf{50}_{-13}$ & $64_{+28}$ & $82.9_{-2}$ \\
          & LlaVA 1.6 & $45_{-32}$ & $39_{-34}$ & $21_{+13}$ & $35_{+20}$ & $\textbf{75}_{+39}$ & $64.1_{+3}$  \\
          & Idefics2 & $80_{-20}$ & $37_{-36}$ & $19_{+5}$ & $25_{+4}$ & $36_{+5}$ & $64.1_{-2}$ \\
         \midrule
          \multirow{4}{*}{\rotatebox[origin=l]{90}{\textbf{Trick}}} & GPT 4  & $\textbf{88}_{+56}$ & $20_{-25}$ & $20_{-15}$ & $10_{-15}$ & $16_{+5}$ & $47.4_{-16}$\\
          & Gemini & $59_{+28}$ & $\textbf{64}_{+8}$ & $\textbf{35}_{-7}$ & $\textbf{38}_{-0}$ & $38_{+9}$ & $\textbf{74.6}_{+3}$ \\
          & LLaVA 1.6 & $22_{-49}$ & $14_{-32}$ & $16_{+13}$ & $14_{+12}$ & $\textbf{66}_{+40}$ & $68.3_{+17}$  \\
          & Idefics2  & $61_{-39}$ & $24_{-39}$ & $20_{+15}$ & $10_{+10}$ & $20_{-3}$ & $57.3_{-2}$\\
         \bottomrule
    \end{tabular}
    }
    \vspace{-1mm}
    \caption{Results in the multi-correct setup where each answer option is considered `locally' with the question, that is, individually with only binary outcome possible (correct or incorrect). Cf., Table~\ref{tab:test_n_correct_main} for the results with the global multiple-choice approach. The numbers in the subscript are deltas to the best corresponding result as reported in Table~\ref{tab:test_n_correct_main}.}
    \label{tab:test_n_correct_binary}
    \vspace{-2.5mm}
\end{table}

\begin{table*}[th!]
    \centering
    \resizebox{\textwidth}{!}{%
    \begin{tabular}{llrrrrrrrrrrrrrrrrrr}
    \toprule
          &  & \multicolumn{3}{c}{0} & \multicolumn{3}{c}{1} & \multicolumn{3}{c}{2} & \multicolumn{3}{c}{3} & \multicolumn{3}{c}{4} &  \multicolumn{3}{c}{F1} \\
          & & Gen & Log & json & Gen & Log & json & Gen & Log & json & Gen & Log & json & Gen & Log & json & Gen & Log & json \\
         \cmidrule(lr){3-5}\cmidrule(lr){6-8} \cmidrule(lr){9-11} \cmidrule(lr){12-14} \cmidrule(lr){15-17} \cmidrule(lr){18-20} 
          \multirow{4}{*}{\rotatebox[origin=l]{90}{\textbf{Counting}}}& GPT 4  & $19_{\pm7.3}$ & $36_{\pm16.2}$ & $12_{\pm16.6}$ & $42_{\pm3.7}$ & $1_{\pm1.2}$ & $38_{\pm15.3}$ & $37_{\pm18.2}$ & $16_{14.5}$ & $52_{\pm21.7}$ & $33_{\pm32.1}$ & $40_{\pm32.3}$ & $54_{\pm36.2}$ & $10_{\pm17.3}$& $0_{\pm0.0}$ & $31_{\pm23.6}$ & $58.4_{12.0}$ & $57.8_{6.3}$ & $62.3_{\pm24.4}$ \\
          & Gemini &  $24_{\pm7.3}$ & - & $14_{\pm22.5}$ & $50_{\pm11.1}$ & - & $56_{\pm7.2}$& $39_{\pm12.1}$ & - & $46_{\pm6.0}$ & $10_{\pm8.7}$ & - & $25_{\pm2.3}$ & $14_{\pm22.0}$ & - & $11_{\pm1.2}$ & $66.1_{\pm5.2}$ & -	& $72.8_{\pm4.0}$  \\
          & Llava 1.6 & $0_{\pm0.0}$ & $0_{\pm0.0}$ & $66_{57.2}$ & $29_{\pm4.6}$ & $39_{\pm9.0}$& $9_{\pm16.2}$ & $1_{\pm1.2}$	& $0_{\pm0.0}$ & $3_{\pm4.6}$ & $0_{\pm0.0}$ & $0_{\pm0.0}$ & $0_{\pm0.0}$ & $3_{\pm4.5}$ & $25_{\pm25.5}$ & $0_{\pm0.0}$ & $41.8_{\pm9.5}$ & $55.3_{\pm3.1}$ & $14.6_{\pm25.3}$ \\
          & Idefics 2 & $0_{\pm0.0}$ & $0_{\pm0.0}$ & $100_{\pm0.0}$ & $42_{\pm2.0}$ & $41_{\pm4.2}$ & $0_{\pm0.0}$ & $0_{\pm0.0}$ & $1_{\pm3.5}$ & $0_{\pm0.0}$ & $2_{\pm3.5}$ & $7_{\pm1.2}$ & $0_{\pm0.0}$ & $5_{\pm7.9}$ & $28_{\pm29.5}$ & $0_{\pm0.0}$ & $41.9_{\pm5.7}$ & $60.2_{5.0}$ & $0_{\pm0.0}$\\
         \midrule
          \multirow{4}{*}{\rotatebox[origin=l]{90}{\textbf{Order}}}& GPT 4 & $17_{\pm{12.7}}$ & $23_{\pm{16.0}}$ & $19_{\pm{19.4}}$ & $33_{\pm{5.0}}$ & $2_{\pm{2.0}}$ & $37_{\pm{8.3}}$ & $21_{\pm{3.1}}$ & $4_{\pm{2.0}}$ & $24_{\pm{6.0}}$ & $5_{\pm{1.2}}$ & $24_{\pm{0.0}}$ & $9_{\pm{2.3}}$ & $0_{\pm{0.0}}$ & $1_{\pm{1.2}}$ & $3_{\pm{3.1}}$ & $55_{\pm{3.3}}$ & $60_{\pm{0.9}}$ & $58_{\pm{5.4}}$\\
        & Gemini & $24_{\pm{12.5}}$ & - & $14_{\pm{22.5}}$ & $50_{\pm{11.1}}$ & - & $56_{\pm{7.2}}$ & $39_{\pm{12.1}}$ & - & $46_{\pm{6.0}}$ & $10_{\pm{8.7}}$ & - & $25_{\pm{2.3}}$ & $15_{\pm{22.0}}$ & - & $11_{\pm{1.2}}$ & $66_{\pm{5.2}}$ & - & $73_{\pm{0.4}}$\\
        & LLaVA 1.6 & $0_{\pm{0.0}}$ & $0_{\pm{0.0}}$ & $81_{\pm{28.4}}$ & $31_{\pm{3.1}}$ & $45_{\pm{10.1}}$ & $9_{\pm{16.2}}$ & $0_{\pm{0.0}}$ & $0_{\pm{0.0}}$ & $1_{\pm{2.3}}$ & $0_{\pm{0.0}}$ & $0_{\pm{0.0}}$ & $0_{\pm{0.0}}$ & $1_{\pm{1.2}}$ & $19_{\pm{17.2}}$ & $0_{\pm{0.0}}$ & $39_{\pm{3.1}}$ & $50_{\pm{7.3}}$ & $14_{\pm{22.7}}$\\
        &   Idefics 2 & $0_{\pm{12.7}}$ & $0_{\pm{16.0}}$ & $100_{\pm{19.4}}$ & $52_{\pm{5.0}}$ & $41_{\pm{2.0}}$ & $0_{\pm{8.3}}$ & $3_{\pm{3.1}}$ & $6_{\pm{2.0}}$ & $0_{\pm{6.0}}$ & $0_{\pm{1.2}}$ & $7_{\pm{0.0}}$ & $0_{\pm{2.3}}$ & $0_{\pm{0.0}}$ & $22_{\pm{1.2}}$ & $0_{\pm{3.1}}$ & $47_{\pm{3.3}}$ & $56_{\pm{0.9}}$ & $0_{\pm{5.4}}$\\
         \midrule
         \multirow{4}{*}{\rotatebox[origin=l]{90}{\textbf{Culture}}}& GPT 4 & $54_{\pm{8.2}}$ & $62_{\pm{20.5}}$ & $36_{\pm{15.3}}$ & $56_{\pm{7.2}}$ & $21_{\pm{7.7}}$ & $68_{\pm{3.9}}$ & $43_{\pm{5.8}}$ & $20_{\pm{6.6}}$ & $55_{\pm{8.7}}$ & $36_{\pm{6.6}}$ & $53_{\pm{3.0}}$ & $41_{\pm{9.0}}$ & $19_{\pm{5.3}}$ & $5_{\pm{1.0}}$ & $18_{\pm{7.3}}$ & $71_{\pm{6.9}}$ & $72_{\pm{5.5}}$ & $78_{\pm{3.3}}$\\
        & Gemini & $57_{\pm{23.1}}$ & - & $21_{\pm{14.1}}$ & $60_{\pm{12.1}}$ & - & $76_{\pm{6.8}}$ & $44_{\pm{21.4}}$ & - & $53_{\pm{1.9}}$ & $39_{\pm{12.3}}$ & - & $63_{\pm{3.9}}$ & $28_{\pm{18.4}}$ & - & $53_{\pm{4.8}}$ & $79_{\pm{5.3}}$ & - & $85_{\pm{0.8}}$\\
        & LLaVA 1.6 & $0_{\pm{0.0}}$ & $0_{\pm{0.0}}$ & $77_{\pm{39.6}}$ & $66_{\pm{4.3}}$ & $73_{\pm{8.4}}$ & $8_{\pm{14.4}}$ & $1_{\pm{1.1}}$ & $0_{\pm{0.0}}$ & $6_{\pm{9.8}}$ & $0_{\pm{0.0}}$ & $15_{\pm{20.7}}$ & $0_{\pm{0.0}}$ & $15_{\pm{1.9}}$ & $36_{\pm{27.4}}$ & $0_{\pm{0.0}}$ & $51_{\pm{1.7}}$ & $61_{\pm{8.6}}$ & $15_{\pm{25.4}}$\\
        & Idefics 2 & $11_{\pm{8.2}}$ & $3_{\pm{20.5}}$ & $100_{\pm{15.3}}$ & $73_{\pm{7.2}}$ & $63_{\pm{7.7}}$ & $0_{\pm{3.9}}$ & $6_{\pm{5.8}}$ & $14_{\pm{6.6}}$ & $0_{\pm{8.7}}$ & $4_{\pm{6.6}}$ & $21_{\pm{3.0}}$ & $0_{\pm{9.0}}$ & $7_{\pm{5.3}}$ & $31_{\pm{1.0}}$ & $0_{\pm{7.3}}$ & $51_{\pm{6.9}}$ & $66_{\pm{5.5}}$ & $0_{\pm{3.3}}$\\
         \midrule
          \multirow{4}{*}{\rotatebox[origin=l]{90}{\textbf{Trick}}} & GPT 4 & $22_{\pm{6.7}}$ & $31_{\pm{9.3}}$ & $12_{\pm{5.9}}$ & $38_{\pm{2.0}}$ & $6_{\pm{4.0}}$ & $45_{\pm{11.7}}$ & $35_{\pm{7.0}}$ & $13_{\pm{4.7}}$ & $35_{\pm{2.9}}$ & $13_{\pm{4.6}}$ & $24_{\pm{6.9}}$ & $25_{\pm{10.1}}$ & $9_{\pm{2.3}}$ & $1_{\pm{1.2}}$ & $11_{\pm{11.7}}$ & $63_{\pm{4.7}}$ & $60_{\pm{4.9}}$ & $61_{\pm{2.3}}$\\
        & Gemini & $30_{\pm{31.9}}$ & - & $3_{\pm{4.6}}$ & $46_{\pm{10.4}}$ & - & $56_{\pm{5.3}}$ & $32_{\pm{21.2}}$ & - & $42_{\pm{4.2}}$ & $20_{\pm{9.2}}$ & - & $37_{\pm{4.6}}$ & $12_{\pm{8.7}}$ & - & $29_{\pm{5.0}}$ & $64_{\pm{10.4}}$ & - & $72_{\pm{0.7}}$\\
        & LLaVA 1.6 & $0_{\pm{0.0}}$ & $0_{\pm{0.0}}$ & $71_{\pm{47.8}}$ & $44_{\pm{5.3}}$ & $46_{\pm{5.3}}$ & $10_{\pm{12.5}}$ & $1_{\pm{2.3}}$ & $0_{\pm{0.0}}$ & $3_{\pm{5.7}}$ & $0_{\pm{0.0}}$ & $0_{\pm{0.0}}$ & $0_{\pm{0.0}}$ & $2_{\pm{3.5}}$ & $26_{\pm{22.3}}$ & $0_{\pm{0.0}}$ & $42_{\pm{2.8}}$ & $51_{\pm{7.9}}$ & $15_{\pm{23.3}}$\\
        & Idefics 2 & $1_{\pm{6.7}}$ & $0_{\pm{9.3}}$ & $100_{\pm{5.9}}$ & $63_{\pm{2.0}}$ & $49_{\pm{4.0}}$ & $0_{\pm{11.7}}$ & $5_{\pm{7.0}}$ & $5_{\pm{4.7}}$ & $0_{\pm{2.9}}$ & $3_{\pm{4.6}}$ & $9_{\pm{6.9}}$ & $0_{\pm{10.1}}$ & $2_{\pm{2.3}}$ & $23_{\pm{1.2}}$ & $0_{\pm{11.7}}$ & $48_{\pm{4.7}}$ & $59_{\pm{4.9}}$ & $0_{\pm{2.3}}$\\
         \bottomrule
    \end{tabular}
    }
    \caption{Accuracy scores in the Multi-Correct Setup, showing the proportion of questions for which the models detected all the answers correctly, split over the questions groups based on the number of gold correct answers (from 0 to 4); F1 scores are also reported to account for partially correct answers as well. Standard deviation due to variation of the prompt (over three different prompts, see Appendix~\ref{sec:prompts}) is also reported in smaller font.}
    \label{tab:test_n_correct_detail_prompts}
\end{table*}

\begin{table*}[t!]
    \centering
    
    \begin{subtable}[t]{0.48\textwidth}
   	\resizebox{0.9\textwidth}{!}{
    \begin{tabular}{llcccccc}
    \toprule
         &  & 0 & 1 & 2 & 3 & 4 & F1 \\ 
         \midrule
          \multirow{16}{*}{\rotatebox[origin=l]{90}{\textbf{Counting}}}& GPT-4 & 20 & 42 &37 & 33 & 10 & 58 \\
          & +all& 4 & 39 & 59 & 60 & 92 & 78 \\
          & +none & 8 & 39 & 51 & 67 & 35 & 71\\
          & +all+none & 6 & 35 & 49 & 67 & 43 & 71 \\
          \cmidrule(lr){2-8}
          & Gemini & 24 & 50 & 39 & 10 & 15 & 66\\
          & +all& 2 & 45 & 27 & 13 & 100 &  78\\
          & +none & 22 & 51 & 31 & 65 & 84 &  79\\
          & +all+none & 14 & 53 & 41 & 67 & 86 & 81 \\
          \cmidrule(lr){2-8}
          & LLaVA 1.6 & 0 & 29 & 1 & 0 & 3 & 42 \\
          & +all& 0 & 34 & 0 & 0 & 4 & 37 \\
          & +none & 0 & 36 & 0 & 0 & 0 & 36  \\
          & +all+none & 0 & 40 & 0 & 0 & 0 & 36 \\
          \cmidrule(lr){2-8}
          & Idefics 2 & 0 & 42 & 0 & 2 & 5 & 42 \\
          & +all& 0 & 48 & 0 & 0 & 0 & 40\\
          & +none & 2 & 44 & 0 & 0 & 0 & 39 \\
          & +all+none & 2 & 44 & 0 & 0 & 0 & 40\\
         \midrule
          \multirow{16}{*}{\rotatebox[origin=l]{90}{\textbf{Order}}}& GPT-4 & 17 & 33 & 21 & 5 & 0 & 55 \\
          & +all& 4 & 42 & 22 & 16 & 18 & 63 \\
          & +none & 6 & 36 & 22& 12 & 6 & 61 \\
          & +all+none & 4 & 38 & 22 & 10 & 2 & 61 \\
          \cmidrule(lr){2-8}
          & Gemini & 24 & 50 & 39 & 10 & 15 & 66\\ 
          & +all& 14 & 58 & 46 & 20 & 36 & 75\\
          & +none & 64 & 46 & 54 & 18 & 4 & 71\\
          & +all+none & 54 & 44 & 52 & 20 & 14 &  72 \\
          \cmidrule(lr){2-8}
          & LLaVA 1.6 & 0 & 31 & 0 & 0 & 1 & 39 \\
          & +all& 0 & 34 & 0 & 0 & 0 & 37 \\
          & +none &  0 & 30 & 0 & 2 & 0 & 38 \\
          & +all+none & 0 & 34 & 0 & 2 & 0 & 38 \\
          \cmidrule(lr){2-8}
          & Idefics 2 & 0 & 52 & 3 & 0 & 0 & 47\\
          & +all& 0 & 56  & 0 & 0 & 8 & 48\\
          & +none & 6 & 58 & 0 & 0 & 0 & 44 \\
          & +all+none & 2 & 58 & 0 & 0 & 8 & 48\\
         \bottomrule
        \end{tabular}
        }
        \end{subtable}
        \begin{subtable}[t]{0.48\textwidth}
   	    \resizebox{0.9\textwidth}{!}{
        \begin{tabular}{llcccccc}
        \toprule
         &  & 0 & 1 & 2 & 3 & 4  & F1 \\ 
         \midrule
          \multirow{16}{*}{\rotatebox[origin=l]{90}{\textbf{Culture}}} & GPT-4  & 54 & 56 & 43 & 36 & 19 & 71 \\
          & +all& 41 & 67 & 43 & 40 & 45 & 80\\
          & +none & 60 & 69 & 47 & 45 & 20 & 80 \\
          & +all+none &  49 & 65 & 45 & 46 & 30 & 79\\
          \cmidrule(lr){2-8}
          & Gemini & 57 & 60 & 44 & 39 & 28 & 79 \\
          & +all& 51 & 58 & ~49 & ~44 & ~87 &  85\\
          & +none & 82 & 57 & 49 & 52 & 42 & 83\\
          & +all+none & 82 & 62 & 49 & 58 & 53 & 85\\
          \cmidrule(lr){2-8}
          & LLaVA 1.6 & 0 & 66 & 1 & 0 & 15 & 51 \\
          & +all& 0 & 60 & 0 & 0 & 19 & 53 \\
          & +none & 24 & 65 & 0 & 0 & 0 & 45\\
          & +all+none & 27 & 67 & 0 & 0 & 6 & 46 \\
          \cmidrule(lr){2-8}
          & Idefics 2 & 11 & 73 & 6 & 5 & 7 & 51 \\
          & +all& 2 & 81 & 0 & 0 & 40  & 63 \\
          & +none & 35 & 81& 0 & 0 & 0 & 50 \\
          & +all+none & 27 & 79 & 0 & 0 & 26 & 60\\
         \midrule
          \multirow{16}{*}{\rotatebox[origin=l]{90}{\textbf{Trick}}} & GPT-4 & 22 & 38 & 35 & 13 & 9 & 60 \\
          & +all& 12 & 40 & 41 & 18 & 22 & 65\\
          & +none & 16 & 44 & 33 & 12 & 8 & 62\\
          & +all+none & 12 & 46 & 35 & 12& 10 & 63 \\
          \cmidrule(lr){2-8}
          & Gemini & 31 & 46 & 32 & 10 & 12 & 64\\
          & +all & 16 & 50 & 37 & 20 & 60 & 73 \\
          & +none & 53 & 36 & 37 & 24 & 22 & 70 \\
          & +all+none & 55 & 40 & 31 & 24 & 30 & 69 \\
          \cmidrule(lr){2-8}
          & LLaVA 1.6 & 0 & 44 & 1 & 0 & 2 & 42 \\
          & +all& 0 & 38 & 0 & 0 & 8  & 42\\
          & +none & 4 & 42 & 0 & 0 & 0 & 39 \\
          & +all+none & 24 & 46 & 0 & 0 & 6 & 38 \\
          \cmidrule(lr){2-8}
          & Idefics 2 & 1 & 63 & 5 & 3 & 2 & 48\\
          & +all & 0 & 58 & 0 & 0 & 30 & 55\\
          & +none & 13 & 58 & 0 & 0 & 0  & 44\\
          & +all+none & 13 & 56 & 0 & 0 & 24  & 53\\
         \bottomrule 
    \end{tabular}
    }
    \end{subtable}
    \caption{Accuracy and F1 score of the models with \textsc{Out-Gen} on the multi-correct setup compared to adding the additional option 'All of the above' (+all), 'None of the above' (+none), and both (+all+none). This illustrates that explicit options for these cases  can improve performance.}
    \label{tab:n_correct_all_none}
\end{table*}

\end{document}